\documentclass{article}

\usepackage{arxiv}

\usepackage{amsmath}
\usepackage{amsfonts}
\usepackage{graphicx}
\usepackage{hyphenat}
\usepackage{balance} 
\usepackage{multirow}
\usepackage{algorithm}
\usepackage{color}
\usepackage{algorithmic}
\usepackage{bm}
\newcommand{\pop}{\mathcal{P}}
\usepackage{subfigure}
\newcommand{\name}{MA3C}
\usepackage{array}
\usepackage{makecell}
\newcommand{\PreserveBackslash}[1]{\let\temp=\\#1\let\\=\temp}

\newcommand{\add}[1]{\textcolor{black}{#1}}

\newcommand{\correct}[2]{\textcolor{black}{#2}}

\usepackage{booktabs}

\title{Communication-Robust Multi-Agent Learning by Adaptable Auxiliary Multi-Agent Adversary Generation}

\author{%
  Lei Yuan\textsuperscript{\rm 1,2},
    Feng Chen\textsuperscript{\rm 1}, Zhongzhang Zhang\textsuperscript{\rm 1},
     Yang Yu\textsuperscript{\rm 1,2,}\thanks{Corresponding Author}\\
  \textsuperscript{\rm 1} National Key Laboratory for Novel Software Technology, Nanjing University, Nanjing, China \\
  \textsuperscript{\rm 2} Polixir.ai\\
  \texttt{\{yuanl, chenf\}@lamda.nju.edu.cn, \{zzzhang,yuy\}@nju.edu.cn}
}

\date{}

\begin{document}

\maketitle

\begin{abstract}

Communication can promote coordination in cooperative Multi-Agent Reinforcement Learning (MARL). Nowadays, existing works mainly focus on improving the communication efficiency of agents, neglecting that real-world communication is much more challenging as there may exist noise or potential attackers. Thus the robustness of the communication-based policies becomes an emergent and severe issue that needs more exploration. In this paper, we posit that the ego system\footnote{\add{Here ego system means the multi-agent communication system itself. We use the word ego to distinguish it from the generated adversaries.}} trained with auxiliary adversaries may handle this limitation and propose an adaptable method of \textbf{M}ulti-\textbf{A}gent \textbf{A}uxiliary \textbf{A}dversaries Generation for robust \textbf{C}ommunication, 
dubbed MA3C, to obtain a robust communication-based policy. In specific, we introduce a novel message-attacking approach that models the learning of the auxiliary attacker as a cooperative problem under a shared goal to minimize the coordination ability of the ego system, with which every information channel may suffer from distinct message attacks. Furthermore, as naive adversarial training may impede the generalization ability of the ego system, we design an attacker population generation approach based on evolutionary learning. Finally, the ego system is paired with an attacker population and then alternatively trained against the continuously evolving attackers to improve its robustness, meaning that both the ego system and the attackers are adaptable.
Extensive experiments on multiple benchmarks indicate that our proposed MA3C provides comparable or better robustness and generalization ability than other baselines. 
\end{abstract}

\section{Introduction}

Communication plays a crucial role in Multi-Agent Reinforcement Learning (MARL)~\cite{zhu2022survey}, with which agents can share information such as experiences, intentions, or observations among teammates to facilitate the learning process, leading to a better understanding of other agents (or other environmental elements) and better coordination as a result. Previous works mainly concentrate on improving communication efficiency from multiple aspects, either by applying (designing) efficient communication protocols~\cite{i2c,wang2020learning,ijcai2022p82,guan2022efficient}, or combining the nature of multi-agent systems to promote communication~\cite{foerster2016learning,DBLP:conf/iclr/KimPS21, maic}, etc, and have been widely demonstrated to \correct{relieve}{alleviate} the partial observability caused by the nature of the environment or the non-stationary caused by the simultaneous learning of multiple agents in a multi-agent system, achieving remarkable coordination for a wide range of tasks like \correct{SMAC}{StarCraft Multi-Agent Challenge (SMAC)}~\cite{maic}. However, the mainstream communication methods are still difficult to be applied in the real world, as these methods popularly assume the policies are trained and tested in a similar or identical environment, seldom considering the policy drift led by noise or hostile attacks \correct{int the environment}{in the environment}.

\correct{Let's retrospect the numerous success}{Let's review the numerous successes} achieved in modern \correct{MARL}{Reinforcement Learning (RL)}.
Most approaches depend highly on deep neural networks, which are, however, shown to be vulnerable to any adversarial attacks~\cite{chakraborty2018adversarial}, i.e., any slight perturbation in the input may lead to entirely different decision-making of a \correct{DRL}{Deep Reinforcement Learning (DRL)} agent~\cite{moos2022robust}. \add{This poses a significant risk to the application of most DRL algorithms, including MARL communication algorithms, because the noise or hostile attacks in the environment can cause the system to crash. Thus, improving the robustness of the decision system, which means that we hope the system still works well when attacked, is an emergent and serve issue. For the mentioned problem in a single-agent system}, many efficient methods are proposed, including adversarial regularizers designing~\cite{DBLP:conf/nips/0001CX0LBH20, oikarinen2021robust}. They enjoy theoretical robustness guarantee, but with limited robustness ability~\cite{xu2022trustworthy}. On the other hand, other approaches introduce auxiliary adversaries to promote robustness via adversarial training, model the training process from a game theory perspective to gain the worst-case performance guarantee and show high effectiveness in different domains~\cite{pan2019risk, zhang2020robust}. As a consequence,  the MARL community also investigates the robustness of
 a multi-agent system from various aspects, including the uncertainty in local observation~\cite{lin2020robustness}, model function~\cite{zhang2020robust}, action making~\cite{hu2022sparse}, etc. However, the communication process in MARL is much more complex.
 For instance, if we consider a fully connected multi-agent with $N$ agents, there are total $N \times (N-1)$ message channels.
If we train an adversary to attack these channels, the attacker's action space may grow dramatically with the number of agents. Previous works make strong assumptions to \correct{relieve}{alleviate} this problem, such as some default channels suffering from the same message perturbation~\cite{xue2022mis} or only a limited number of agents sustaining some heuristic noise injection. \add{Despite the difficulty of considering the communication robustness of the multi-agent system, it is an important topic because the communication channels are likely to be under noise or hostile attacks in some application scenarios. This current state motivates us to design reasonable approaches to improve the robustness of the communication system when message attacks are possible in the environment.}


In this work, we take a further step towards achieving robustness in MARL communication via auxiliary adversarial training. We posit a robust communication-based policy should be robust to scenarios where \textbf{every} message channel may be perturbed under \textbf{different} degrees at \textbf{any time}. Specifically, we model the message adversary training process as a cooperative MARL problem, where each adversary obtains the local state of one message sender, then outputs $N-1$ stochastic actions as message perturbations for each message to be sent to other teammates. For the optimization of the adversary, as there are $N$ adversaries coordinating to minimize the ego system's return, we can use any cooperative MARL approach
to train the attacker system. Moreover, to \correct{relieve}{alleviate} the overfitting problem of using a single attacker\add{~\cite{vinitsky2020robust}}, we introduce an attacker population learning paradigm, with which we can obtain a set of attackers with high attacking quality and behavior diversity. The ego system and the attacker are then trained in an alternative way to obtain a robust communication-based policy. Extensive experiments are conducted on various cooperative multi-agent benchmarks that need communication to coordination, including Hallway~\cite{ndq},  two maps from StarCraft Multi-Agent Challenge (SMAC)~\cite{ndq}, a newly created environment Gold Panner (GP), and Traffic Junction~\cite{tarmac}. The experimental results show that MA3C outperforms multiple baselines. Further, more results validate it from other aspects, like generalization and high transfer ability for complex tasks. 


\section{Related Work}
\textbf{Multi-Agent Communication.} Communication is a significant topic in MARL under partial observability, which typically studies \textbf{when} to send \textbf{what} messages to \textbf{whom}~\cite{zhu2022survey}. The early relevant works mainly consider combining communication with any existing MARL methods, using broadcasted messages to promote coordination within a team~\cite{foerster2016learning} or designing end-to-end training paradigms that 
update the message network and policy network together with the back-propagated gradients~\cite{communication16}.
To improve communication in complex scenarios, researchers investigate the efficiency of communication from multiple aspects like designing positive listening protocol~\cite{DBLP:conf/atal/LoweFBPD19, DBLP:conf/nips/EcclesBLLG19}.
 To avoid redundant communication, 
 some works employ techniques such as gate mechanisms~\cite{acml,i2c,ijcai2022p82} to decide whom to communicate with, or attention mechanisms~\cite{tarmac,doubleattention,wang2021tom2c,guan2022efficient} \correct{to exact}{to extract} the most valuable part from multiple received messages for decision-making.
 What content to share is also a critical point. A direct practice is to only share local observations or their embeddings~\cite{foerster2016learning, ndq}, but it inevitably causes bandwidth wasting or even degrades the coordination efficiency.
 Some methods utilize techniques like teammate modeling to generate more succinct and efficient messages~\cite {vbc,tmc,maic}. Besides, VBC~\cite{vbc} and TMC~\cite{tmc} also answer the question of when to communicate by utilizing a fixed threshold to control the chance of communication.
 In terms of the robustness of communication in cooperative MARL, \cite{mitchell2020gaussian} filters valuable content from noisy messages by Gaussian process modeling. AME~\cite{sun2022certifiably} utilizes an ensemble-based defense method to reach robustness but it only assumes no more than half of the message channels in the system can be attacked. \cite{xue2022mis} considers the communication robustness in situations where one agent in the cooperating group is taken over by a learned adversary, and then the policy-search response-oracle (PSRO) technique is applied to achieve communication robustness.
 

\textbf{Robustness in Cooperative MARL.} Previous cooperative MARL~\cite{oroojlooyjadid2019review} works either concentrate on improving coordination ability from diverse aspects like scalability~\cite{christianos2021scaling}, credit assignment~\cite{wang2021towards}, and non-stationarity~\cite{papoudakis2019dealing}, or applying the cooperative MARL technique to multiple domains like autonomous vehicle teams~\cite{peng2021learning,kouzehgar2020multi}, power management~\cite{wang2021multi}, and dynamic algorithm configuration~\cite{xue2022multiagent}. Those approaches ignore the robustness of the learned policy when encountering uncertainties, perturbations, or structural changes in the environment~\cite{moos2022robust}, hastening the robustness test in the MARL~\cite{guo2022towards}. For the altering of the opponent policy, M3DDPG~\cite{li2019robust} learns a minimax variant of MADDPG~\cite{maddpg} and trains the MARL policy in an adversarial way, showing potential in solving the problem of poor local optima compared with multiple baselines. \cite{van2020robust} applies the social empowerment technique to avoid the MARL overfitting to their specific trained partners.
As for the uncertainty caused by the inaccurate knowledge of the MARL dynamic model, R-MADDPG~\cite{zhang2020robustmarl} proposes the concept of robust Nash equilibrium, treats the uncertainty of environment as a natural agent, and exhibits superiority when encountering reward uncertainty. Consider the observation perturbation in cooperative MARL, \cite{lin2020robustness} learns an adversarial observation policy to attack one participant in a cooperative MARL system, demonstrating the high vulnerability of cooperative MARL facing observation perturbation. For the action robustness in cooperative MARL,  ARTS~\cite{phan2020learning} and RADAR~\cite{phan2021resilient} learn resilient MARL policies via adversarial value decomposition.~\cite{hu2022sparse} further designs an action regularizer to attack the cooperative MARL system efficiently. 


 \textbf{Population-Based Reinforcement Learning (PBRL).} Population-Based Training (PBT) has been widely used in machine learning and made tremendous success in different domains~\cite{jaderberg2017population}, which also reveals great potential for reinforcement learning problems~\cite{jaderberg2019human,DBLP:journals/fcsc/QianY21}. One successful application of PBRL is to train multiple policies to generate diverse behaviors that can accelerate the learning of downstream tasks~\cite{derek2021adaptable}. Another category focus on applying population training to facilitate reinforcement learning in aspects like efficient exploration~\cite{parker2020effective}, model learning~\cite{luo2022survey}, robustness~\cite{vinitsky2020robust}, and zero-shot coordination~\cite{zhao2021maximum,xue2022heterogeneous}. 
 \add{Among all these works, \cite{vinitsky2020robust} is most similar to our work, which maintains a population with different individuals by the different network initialization, without an explicit diversity constraint among individuals. However, our work differs because we further consider the robustness of multi-agent communication beyond the single-agent RL setting, and we explicitly optimize the diversity of the attacker population. Actually, if all individuals act indistinctively, the population is just multiple copies of an individual, so one key factor in population-based training is to evaluate individuals effectively and ensure their differences.}
One series of representative algorithms is evolutionary computation~\cite{DBLP:conf/nips/Parker-HolderPC20,wang2021evolutionary}, and among them the Quality-Diversity (QD) algorithms~\cite{qd, qd-survey-optimization} have been widely used to obtain high-quality solutions with diverse behaviors in a single run. High-quality refers to each individual trying to accomplish the given task, while diversity means all individuals behave differently as possible. These methods obtain great success in diverse domains such as skill learning~\cite{lim2022dynamics}, multi-object optimization~\cite{DBLP:conf/gecco/PierrotRBC22}, skill discovering~\cite{chalumeau2022neuroevolution}, etc. Nevertheless, multi-agent problems such as SMAC~\cite{smac} are much more complex, as how to distinguish agents' cooperation patterns and measure distances between different joint policies are still open questions. We will apply an efficient mechanism to solve it.

\section{Problem Setting}\label{sec:problem_setting}
This paper considers the problem setting of fully cooperative MARL which can be modeled as a Decentralized Partially Observable Markov Decision Process (Dec-POMDP)~\cite{oliehoek2016concise} consisting of a tuple $\langle \mathcal{I}, \mathcal{S}, \mathcal{A}, P, \Omega, O, R, \gamma\rangle$, where $\mathcal{I}=\{1,\dots,N\}$ indicates the finite set of $N$ agents and $\gamma\in[0,1)$ is the discounted factor. At each timestep, each agent $i$ observes a local observation $o_i\in \Omega$, which is a projection of the true state $s\in S$ by the observation function $o_i=O\left(s,i\right)$. Each agent selects an action $a_i\in \mathcal{A}$ to execute, and all individual actions form a joint action $\bm{a}\in\mathcal{A}^N$ which leads to the next state $s'\sim P\left(s'|s,\bm{a}\right)$ and a reward $r=R\left(s,\bm{a}\right)$. Besides, a message set $\mathcal{M}$ is introduced to model the agent communication, and the Dec-POMDP can be transformed into a Dec-POMDP with Communication (Dec-POMDP-Comm)~\cite{xue2022mis}. Specifically, each agent makes decision based on an individual policy $\pi_i\left(a_i\mid \tau_i, m_i\right)$, where $\tau_i$ represents the history $\left(o_i^1,a_i^1,\dots,o_i^{t-1},a_i^{t-1},o_i^t\right)$ of agent $i$, $m_i\in\mathcal{M}$ is the message received by agent $i$ and $m_{ij}$ indicates the message sent from agent $j$ to $i$. As each agent can behave as a message sender and receiver, this paper uses the default message generator and message processor for various methods like NDQ~\cite{ndq}. Specifically, the messages sent by agent $i$ are denoted as $m_{:i}=msg_i(o_i)$, where $msg_i(\cdot)$ indicates the message generator of agent $i$.
It focuses on obtaining a robust policy via adversarial training for different message perturbations.

To conduct the adversarial training, we aim to learn auxiliary message adversaries $\hat{\pi}$ which perturb the messages received by each agent, transforming $m$ into $\hat{m}$. Thus, each agent $i$ actually takes action by $\pi_i(a_i|\tau_i, \hat{m}_i)$. Specifically, we apply additive perturbations and $\hat{\pi}$ is defined as a deterministic policy, which means that:
\begin{equation}
    \bm{\xi} =\hat{\pi}(\correct{s}{\bm{o}}),~\hat{m} = m + \bm{\xi},
\end{equation}
\correct{where $\hat{\pi}$ is a centralized policy that takes the state as input. In practice, we also consider decentralized attack policy, where $\hat{\pi}=\left(\hat{\pi}_1,\dots,\hat{\pi}_N\right)$, which will be detailed in Sec.~\ref{sec:maattack}.}{where $\bm{o}$ is the joint observation of all agents, i.e., $\bm{o}=(o_1,o_2,\cdots,o_n)$. In practice, we consider decentralized attack policy, where $\hat{\pi}=\left(\hat{\pi}_1,\dots,\hat{\pi}_N\right)$ with each sub-policy taking the local observation $o_i$ as input. More details are described in Sec.~\ref{sec:maattack}.}
Besides, arbitrary perturbations without bounds can have a devastating impact on the communication performance, and robustness under this situation is almost impossible. Hence to consider a more realistic setting, we restrict the power of adversaries and constrain the perturbed messages to a set $\mathcal{B}$. For example, we can typically define $\mathcal{B}$ as a $p$-norm ball centered around the original messages, i.e., $\mathcal{B}=\{\hat{m}\mid\|m-\hat{m}\|_p = \epsilon\}$, where $\epsilon$ is the given perturbation magnitude and $p$ is the norm type.

Finally, we optimize the ego-system policy by value-based MARL method, where deep Q-learning \cite{mnih2015human} implements the action-value function  with a deep neural network $Q\left(\boldsymbol\tau, \boldsymbol a; \psi\right)$ parameterized by $\psi$. This paper follows the Centralized Training and Decentralized Execution (CTDE)~\cite{gronauer2022multi} paradigm. In the centralized training phase, it uses a replay memory $\mathcal{D}$ to store the transition tuple $\langle \boldsymbol{\tau}, \boldsymbol{a}, r, \boldsymbol{\tau}^\prime \rangle$. We use $Q\left(\boldsymbol\tau, \boldsymbol a; \psi\right)$ to approximate $Q\left(s, \boldsymbol a\right)$ to \correct{relieve}{alleviate} the partial observability. Thus, the parameters $\psi$ are learnt by minimizing the expected Temporal Difference (TD) error:
  \begin{equation}
     \mathcal{L}(\psi)=\mathbb{E}_{\left(\boldsymbol{\tau}, \boldsymbol{a}, r, \boldsymbol{\tau}^{\prime}\right) \in \mathcal{D}}\left[\left(r+\gamma V\left(\boldsymbol{\tau}^{\prime} ; \psi^{-}\right)-Q(\boldsymbol{\tau}, \boldsymbol{a} ; \psi)\right)^{2}\right],
 \end{equation}
 where $V\left(\boldsymbol{\tau}^{\prime} ; \psi^{-}\right)=\max _{\boldsymbol{a}^{\prime}} Q\left(\boldsymbol{\tau}^{\prime}, \boldsymbol{a}^{\prime} ; \psi^{-}\right)$ is the expected future return of the TD target and $\psi^-$ are parameters of the target network, which will be periodically updated with $\psi$. \add{Note that though we follow the CTDE paradigm, multi-agent communication is allowed in the execution process, which means that the inputs to the individual Q-networks are agents' local observations and the received information.}
 More specific details about the design of $Q\left(\bm{\tau},\bm{a};\psi\right)$ are related to the corresponding underlying multi-agent communication methods. \correct{Such as in Full-Comm}{Such as in the Full-Comm algorithm~\cite{guan2022efficient} where agents broadcast their individual observations}, $Q\left(\bm{\tau},\bm{a};\psi\right)$ is decomposed of a mixing network and $N$ individual Q-networks, where each $Q_i$ is conditioned on both agent $i$'s observation and received messages.

\section{Method}
\label{sec:method}
In this section, we will describe the detailed design of our proposed method named \name. Firstly, we show how to model message adversaries as a cooperative multi-agent attack problem and how to learn a specific attacker instance. Immediately after, we introduce the concept of attacker population and our diversity mechanism that helps obtain an auxiliary attacker population with diverse and qualified attacker instances. Finally, a complete training description of our approach is provided. 
\begin{figure}[t]
    \centering
    \includegraphics[width=1.0\linewidth]{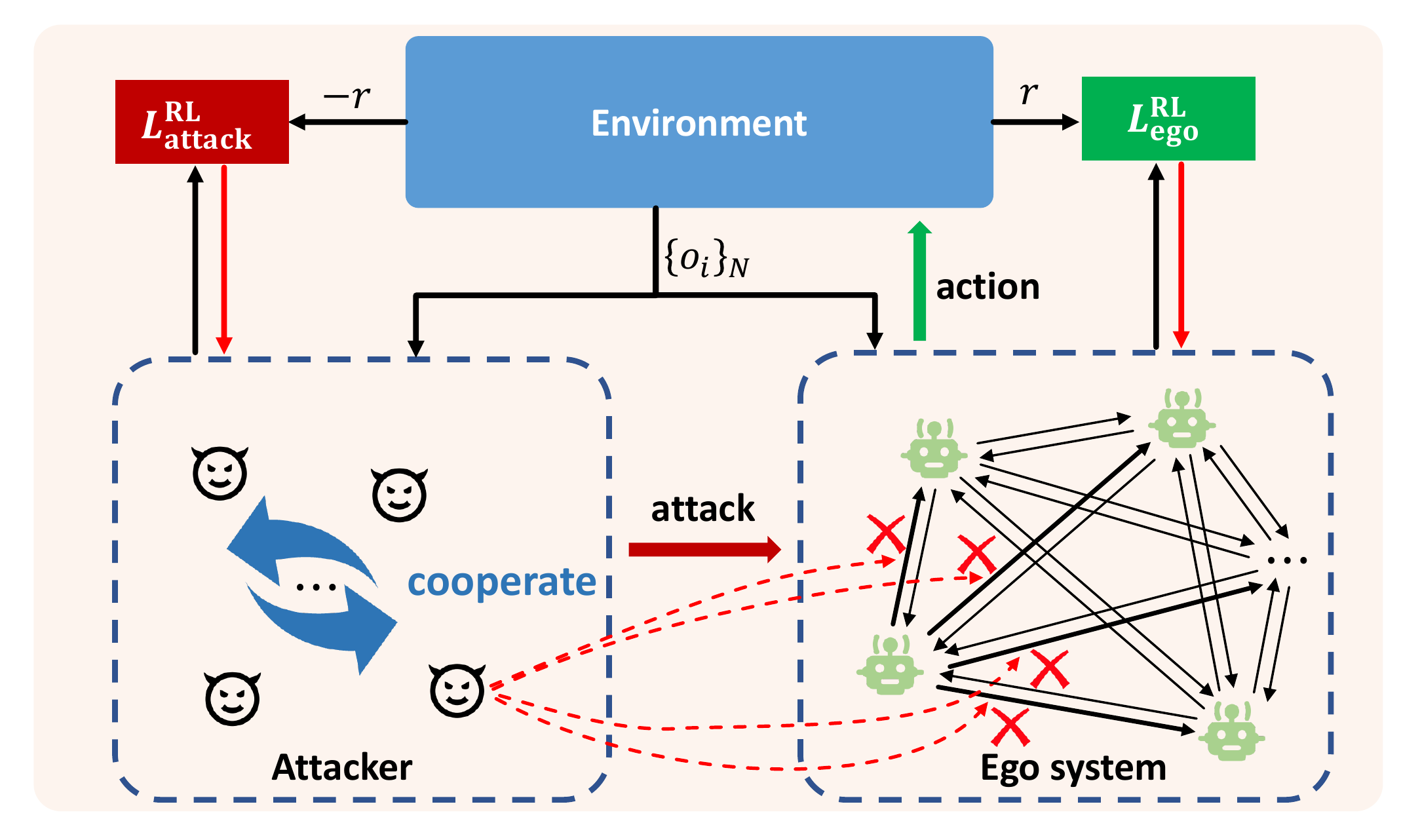}
    \caption{The overall relationship between the attacker and the ego system. The black solid arrows indicate the direction of data flow, the red solid ones indicate the direction of gradient flow and the red dotted ones mean the attack actions from the attacker onto specific communication channels.}
    \label{fig:information_attacker}
\end{figure}
\subsection{Message Channel Level Attacking}
\label{sec:maattack}
To achieve robust communication for a multi-agent system, we propose to train the ego system policy under possible communication channel attacks, thus consequently obtaining an attack-robust communication-based policy.
Generally, we aim to design an adversarial framework that consists of two main-body components, which are respectively the \textbf{attacker} and \textbf{ego system} (c.f. Fig.~\ref{fig:information_attacker}). Specifically, the attacker aims to produce perturbation on the communication channels to degrade the communication performance of the ego system. In contrast, the ego system can be any MARL methods with communication, aiming at maximizing coordination ability. The core idea is that adversarial training helps the ego system encounter different message perturbation situations and learn how to handle possible communication attacks to achieve robust communication.
To serve this goal, one critical question is how to build a qualified attacker. 

To answer the mentioned question, we claim that a qualified attacker ought to have two vital properties - \textit{\textbf{comprehensiveness}} and \textit{\textbf{specificity}}. Mostly, every agent plays a different role in a multi-agent system, and each communication channel is of different importance to the whole communication system. In most cases, a portion of the message channels are more important, and perturbing these message channels is the key to achieving a practical attack. The referred \textit{\textbf{comprehensiveness}} means that the attacker should consider all communication message channels to avoid only perturbing some specific or even useless messages, but overlooking those vital individuals, leading to low robustness. On the other hand, since different agents process the received messages in different ways in general, a reasonable perspective is that the attacker should be distinct in how it perturbs the messages received by each agent, which we call \textit{\textbf{specificity}}.

Except for these two important properties, one more fundamental requirement for the attacker is that it should act sequentially to degrade the performance of the ego system. 
To serve this goal, we model the attacking process as a sequential decision problem. Specifically, the reward is defined as the opposite of the ego system's reward: $\hat{R} = -R$, and the learning objective is to maximize:
\begin{equation}
\begin{aligned}
\mathbb{E}\left[\sum_{t=0}^\infty \gamma^t \hat{R}\right]=\mathbb{E}\Bigg[-\sum_{t=0}^\infty&\gamma^t R\left(s^t,\hat{\bm{a}}^t\right) \bigg|s^{t+1}\sim P\left(\cdot|s^t, \hat{\bm{a}}^t\right),\\
&\hat{a}_i^t\sim \correct{\hat{\pi}}{\pi_i}\left(\cdot\mid o_i^t, \hat{m}_i^t\right), \hat{m}_i^t=m_i^t+\xi_i^t,\\
&m^t_{:i} = msg_i\left(o_i^t\right), \bm{\xi}^t = \hat{\pi}(s^t)\Bigg].
\end{aligned}
\end{equation}
Note that to equip the attacker with the properties of \textit{\textbf{comprehensiveness}} and \textit{\textbf{specificity}}, we model the action space as 
a continuous action space
with dimension of $N\times (N-1)\times d_{comm}$, where $N$ denotes the number of agents, $N\times(N-1)$ is the total number of the communication channels and $d_{comm}$ indicates the dimension of one single message.
At each time step, the $attacker$ should learn to output an action with dimension of $N\times (N-1) \times d_{comm}$ which serves as the perturbation on the communication messages:
\begin{equation}
\begin{aligned}
    \bm{\xi}^t &= \hat{\pi}(s^t),~\bm{\xi}^t\in \mathbb{R}^{N\times(N-1)\times d_{comm}},\\
    \hat{m}_{ij}^t &= m_{ij}^t + \xi_{ij}^t,~i\neq j\in \{1, 2, \cdots, N\}.
\end{aligned}
\end{equation}
However, this design leads to a large action space, especially when there exist quite many agents in the environment since the action dimension grows squarely with the number of agents. The problem of large action space is likely to bring difficulties to the attack policy learning, which has already been discussed before~\cite{christianos2021scaling, zhang2021multi}. In fact, some modern MARL algorithms like QMIX~\cite{
qmix}, decompose the joint action space and alleviate the difficulties posed by the large joint action space in policy learning.
Motivated by this point, we propose to apply cooperative MARL methods to mitigate the problem of large action space in attack policy learning.

Specifically, we construct $N$ virtual agents, divide the $N\times(N-1)$ communication channels into $N$ groups and let each virtual agent be responsible for the attacks on $N-1$ communication channels. In other words, the action dimension for each virtual agent is $(N-1)*d_{comm}$, which grows linearly with the number of agents, and these $N$ virtual agents together form a concept of \textbf{attacker}. As shown in Fig.~\ref{fig:information_attacker}, the $i$-th virtual agent takes the individual observation of agent $i$ as the input into its policy. The policy's output is the perturbation on the sent information from agent $i$, i.e., virtual agent $i$ undertakes the responsibility of contaminating the messages sent by the $i$-th agent. In this way, we can think that these $N$ virtual agents cooperate to attack the underlying communication system effectively. The whole process of adversarial training can be seen as a confrontation between two groups of agents; one is the group of these $N$ virtual agents, and the other is the underlying multi-agent system (the ego system). In particular, our approach can be applied to any off-the-self MARL algorithm that can handle problems with continuous action spaces.

In practice, adversary $i$ (the virtual agent mentioned above) uses the current observation $o_i$ of agent $i$, then learns an attack policy $\hat{\pi}_i(\xi_{i}|o_i; \theta)$ to perturb the $N-1$ messages sent by agent $i$, where $\theta$ denotes the parameters of the total attacker model. Thus each message equals to $m_{ij}+\xi_{ij}$. We here can employ any actor-critic method to optimize the policy. However, in the MARL system, there exists only a shared reward; thus, the simplest way is to use a central critic similar to COMA~\cite{DBLP:conf/aaai/FoersterFANW18} to optimize it. It takes $(s, \bm{\xi})$ as input and outputs the Q value $Q(s, \bm{\xi})$. Specifically, we extend TD3~\cite{DBLP:conf/icml/FujimotoHM18} to multi-agent setting, named MATD3, where the centralized Q-function is learned with
\begin{equation}
    \begin{aligned}
        &\mathop{\arg\min}_{\theta} \sum_{j=1}^2 \add{\sum_{t}} \left(Q_j\left(s^t, \xi^t_1, \dots, \xi^t_N\right) - y^t\right)^2, ~ j\in\{1, 2\} \\
        &y^t = \mathbb{E}\bigg[ r^t + \gamma \min_{j=1,2}Q_j'\left(s^t, \hat{\pi}_1(o^{t+1}_i), \dots, \hat{\pi}_N(o^{t+1}_N)\right)\bigg],
    \end{aligned}
\end{equation}
where we maintain two Q-networks $Q_1, Q_2$, and $Q_j'$ is the target network for $Q_j$. The actors are optimized via the deterministic policy gradient:
\begin{equation}
    \begin{aligned}
    \nabla_\theta J = \mathbb{E}\left[\nabla_\theta \hat{\pi}_i\left(\xi_i|o_i\right)\nabla_{\xi_i}Q_1\left(s, \xi_1, \dots, \xi_N\right)\right].
    \end{aligned}
    \label{loss_qmix}
\end{equation}

\subsection{Attacker Population Optimization} \label{populationoptimization}
\begin{figure*}[t]
    \centering
    \includegraphics[width=1.0\linewidth]{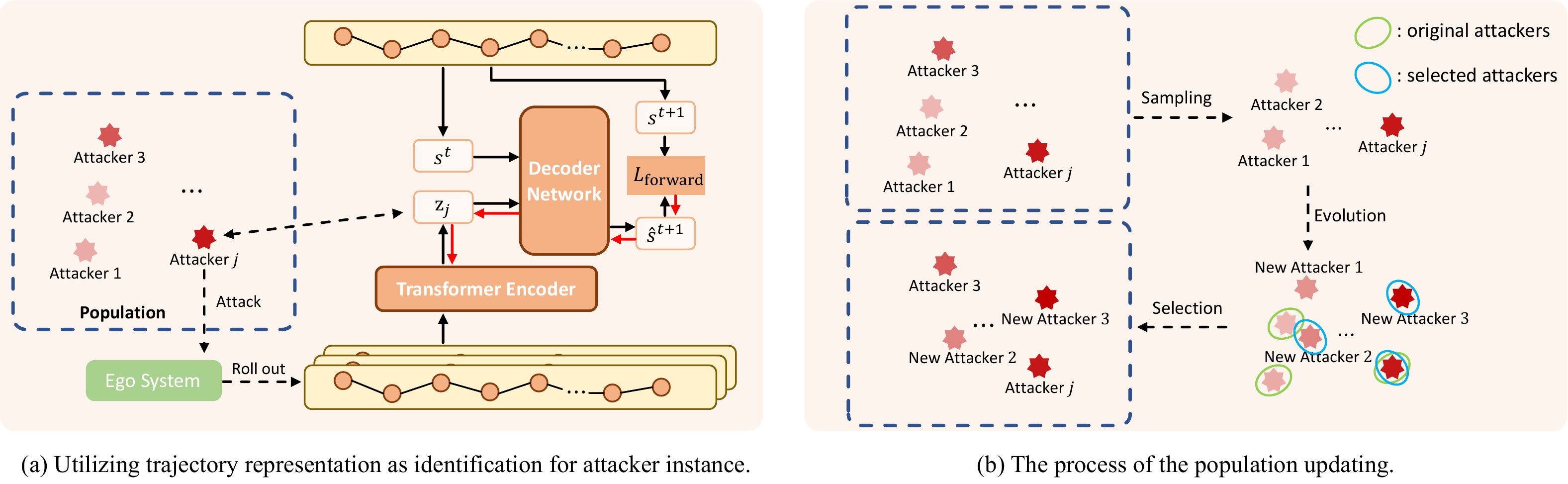}
    \caption{The overall framework for the attacker population optimization. (a) We utilize the representation of the attacked ego system's trajectories to identify different attacker instances. Specifically, we apply an encoder-decoder architecture to learn the trajectory representation. The black solid arrows indicate the direction of data flow and the red solid ones imply the direction of gradient flow. (b) This is a simple visualization case for one time population updating. The locations of points imply the distances of representations and the color shades indicate the attack ability, i.e., the attackers corresponding to deeper points are stronger attackers. For example, new attacker $3$ is accepted as it is distant enough with other attackers, and the oldest attacker $1$ is removed; new attacker $2$ is accepted and the closest attacker $2$ is removed as it is weaker.}
    \label{fig:attacker_population}
\end{figure*}  
Many previous adversarial algorithms for reinforcement learning typically build one attacker and alternatively do adversarial training via finding a Nash equilibrium solution. However, one often criticized issue is that the ego system is largely over-fitted to the corresponding attacker and may fail to obtain good performance in practice. One possible solution to this issue is constructing an attacker population instead of one single attacker. 
By forcing the ego system to perform well against all attackers in the population, we expect to impede the ego system from over-fitting to one specific pattern and help it achieve good performance in the average sense. We believe this practice can effectively help avoid some crash cases, which is consistent with the goal of robustness.

Besides, one natural idea is that the adversarial training population should be diverse enough to cover attackers with quite different patterns, thus avoiding the population degenerating into one single attacker. This requires that we train the ego system along with the learned attacker population as:
\begin{equation}
\begin{aligned}
    &\mathop{\arg\max}_{\pi} J\left(\pi|\pop\right) = \frac{1}{|\pop|}\sum_{\hat{\pi}\in \pop} J\left(\pi|\hat{\pi}\right),\\
    &\begin{aligned}
    J\left(\pi|\hat{\pi}\right)=\mathbb{E}\Bigg[-&\sum_{t=0}^\infty\gamma^t R\left(s^t,\hat{\bm{a}}^t\right) \bigg|s^{t+1}\sim P\left(\cdot|s^t, \hat{\bm{a}}^t\right),\\
    &\hat{a}_i^t\sim \correct{\hat{\pi}}{\pi_i}\left(\cdot\mid o_i^t, \hat{m}_i^t\right), \hat{m}_i^t=m_i^t+\xi_i^t,\\
    &m^t_{:i} = msg_i\left(o_i^t\right), \bm{\xi}^t = \hat{\pi}(s^t)\Bigg],
    \end{aligned}\\
    &\mathtt{Distance}(\hat{\pi}_i,\hat{\pi}_j) > \epsilon,~\forall \hat{\pi}_i,\hat{\pi}_j \in \pop,\label{equ:distance}
\end{aligned}
\end{equation}
where $\hat{\pi}_j \in \pop$ means we sample an attacker from the learned population during the training process, and $\mathtt{Distance}(\hat{\pi}_i,\hat{\pi}_j)$ refers to the distance between attacker $i$ and attacker $j$. Nevertheless, the remaining question is how to define the $\mathtt{Distance}$ function.
Some previous population-based methods~\cite{DBLP:conf/nips/Parker-HolderPC20,wang2021evolutionary} typically achieve a balance between the quality and diversity by utilizing the mean action vector as an identification for each instance. However, these practices usually require a well-defined behavioral descriptor~\cite{cully2019autonomous}, which is sometimes the prior knowledge about some critical states and has demonstrated even hard in MARL setting~\cite{zhou2022continuously}. 


To reach the mentioned goal, firstly, we observe that the differences in attack patterns can be well revealed in the behaviors of the attacked ego system. The basic idea is that when under different types of communication attacks, the multi-agent system may exhibit different behaviors, e.g., generating quite different trajectories.
Based on this idea, for each attacker $j$, we sample $m$ trajectories of the ego system under the communication attacks of attacker $j$, and utilize a trajectory encoder to encode them into a representation $z_j$: 
\begin{equation}
    z_j = \frac{1}{m}\sum_{k=1}^m f^{\rm{enc}}_{\phi}(\tau_j^k),~
\end{equation}
where $f^{\rm{enc}}_\phi$ denotes a trajectory encoder network which is parameterized by $\phi$. Then we utilize the representation $z_j$ as the identification for attacker $j$, and the distance between $z_i$ and $z_j$ describes how different these two attackers are.

\begin{algorithm}[t!]
    \caption{\texttt{Population Update}}
    \label{alg:pop_update}
    \renewcommand{\algorithmicrequire}{\textbf{Input:}}
    \renewcommand{\algorithmicensure}{\textbf{Output:}}
    
    \begin{algorithmic}[1]
        \REQUIRE Population $\mathcal{P}$, Distance threshold $\zeta$, Number of evolution $T_{\rm{evolution}}$, Ego system model $\pi_{\rm{ego}}$, Trajectory encoder $f^{\rm{enc}}_{\theta}$.\\
        
        \FOR{$t=1, 2, \dots, T$}
            \STATE Select a subset $\pop_{\rm{sub}}$ of a fixed size from $\pop$ randomly.
            \FOR{attacker instance $I_{\rm{attacker}}$ in $\pop_{\rm{sub}}$}
                \STATE Update $I_{\rm{attacker}}$ with the MATD3 algorithm and obtain a new attacker instance $I_{\rm{attacker}}'$.
                \STATE Roll out $m$ trajectories $\{\tau\}_m$ of the ego system $\pi_{\rm{ego}}$ under the attack of $I_{\rm{attacker}}'$.
                \STATE Encode the trajectories $\{\tau\}_m$ with $f^{\rm{enc}}_{\theta}$ and get identification $z'$ for the new attacker instance $I_{\rm{attacker}}'$.
                \STATE Select the attacker instance $\bar{I}_{\rm{attacker}}$ from $\pop$, which has the closest identification $\bar{z}$ to $z'$.
                \IF{$\|\bar{z} - z'\|_2 > \zeta$}
                    \STATE Remove the oldest attacker instance in $\pop$.
                    \STATE Add $I_{\rm{attacker}}'$ to $\pop$.
                \ELSE
                    \IF{$I_{\rm{attacker}}'$ is stronger than $\bar{I}_{\rm{attacker}}$}
                        \STATE Remove $\bar{I}_{\rm{attacker}}$ from $\pop$.
                        \STATE Add $I_{\rm{attacker}}'$ to $\pop$.
                    \ELSE
                        \STATE Discard $I_{\rm{attacker}}'$.
                    \ENDIF
                \ENDIF
            \ENDFOR
        \ENDFOR
    \end{algorithmic}
\end{algorithm}

In designing the architecture and the optimization objective of the trajectory encoder network, we emphasize two vital points: (i) some specific parts in one trajectory are essential for distinguishing it from other trajectories; (ii) the representation $z_j$ should imply the behavioral trends of the ego system when under the attacks of attacker $j$. To address point (i), we design the architecture of the trajectory encoder as a transformer~\cite{DBLP:conf/nips/VaswaniSPUJGKP17} network because the attention mechanism can help the encoder focus on specific essential parts of the trajectory. In terms of point (ii), we design a forward prediction loss $L_{\rm{forward}}$ to help optimize the trajectory encoder network, as shown in Fig.~\ref{fig:attacker_population}(a). Specifically, we feed $z_j$ and $s_j^t$ into an extra decoder network, which outputs a prediction for $s_j^{t+1}$. We update the encoder and decoder networks together by minimizing the following prediction loss:
\begin{equation}
\begin{aligned}
    L_{\rm{forward}}(\phi, \eta) = \sum_{k=1}^{m} \sum_{s_j^{t+1}\sim\tau_j^k}\|s_j^{t+1} - \hat{s}_j^{t+1} \|^2,\\
    \hat{s}_j^{t+1} = f^{\rm{dec}}_\eta(z_j, s_j^t),~t\in \{ 0,1,2,\cdots,T-1\},
\end{aligned}
\end{equation}
where $s_j^{t+1}$ denotes the state information at timestep $t+1$ in one sampled trajectory, and $f^{\rm{dec}}_\eta$ indicates the decoder network parameterized by $\eta$. The key point is that by optimizing the forward prediction loss, we force the encoded representation $z_j$ to contain behavioral information which can help predict the trend of these sampled trajectories. 

After defining the identification for attackers, we optimize the attacker population by alternatively conducting the processes of evolution and selection. For evolution, we update the selected instances via the method described in Sec.~\ref{sec:maattack}, leading to some new instances.
For selection, we always choose those more distant new instances to add to the population based on the definition of $z_j$.
In specific, each time we want to update the population, we randomly select a fixed proportion of attacker models from the population.
We apply MATD3 (c.f. Sec.~\ref{sec:maattack}) for each selected instance to update its attack policy with a fixed number of samples, resulting in a group of new attacker instances. Then for each new attacker, we find the closest instance to it in the population and compare their distance based on the trajectory representation. If their distance is over a fixed threshold,  we retain the new attacker and throw out the current \textit{oldest instance} in the population; otherwise, we retain the better one by comparing their attack performance and throw out the other. Note that we utilize a First-In-First-Out (FIFO) queue to implement the attacker population and the \textit{oldest instance} here means the first element of the queue. The whole process 
is shown in Alg.~\ref{alg:pop_update}.
\begin{algorithm}[t!]
    \caption{\texttt{Adversarial Training}}
    \label{alg:adv_train}
    \renewcommand{\algorithmicrequire}{\textbf{Input:}}
    \renewcommand{\algorithmicensure}{\textbf{Output:}}
    
    \begin{algorithmic}[1]
        \REQUIRE Initialized population $\pop$, Initialized ego system policy $\pi_{\rm{ego}}$.
        \STATE Pre-train the ego system without any adversaries.
        \FOR{iter = 1, 2, $\dots$, max\_iter}
            \STATE Call \texttt{Population Update} to update the attacker population $\pop$.
            \STATE Update the ego system policy $\pi_{\rm{ego}}$ with a multi-agent communication method against the whole attacker population $\pop$.
        \ENDFOR
    \end{algorithmic}
\end{algorithm}

\subsection{Robust Communication and Training}
Based on the proposed attacker training algorithm and the population optimization method, we further design a whole training framework where we alternatively train the ego system and update the population. In fact, throughout the whole process of adversarial training, we maintain an ego system and a fixed-size population.
In the phase of ego-system training, we uniformly select an attacker from the population at the start of each episode. We let the ego system roll out with this attacker, and the roll-outed trajectory is added to a training buffer. We update the ego system with data sampled from the buffer, which equates to adversarially training the ego system against the whole attacker population.
On the other hand, in the phase of population updating, we load the latest ego system model and apply the population updating mechanism against the loaded ego system, as described in Alg.~\ref{alg:pop_update}. The whole training process consists of multiple repetitions of these two phases, and the ego system and the population are iteratively enhanced in the whole process. The whole adversarial training procedure is described in Alg.~\ref{alg:adv_train}.

\section{Experimental Results}

Note that our approach \correct{is vertical to}{is orthogonal to} the underlying multi-agent communication algorithm, thus to validate the effectiveness of our approach in this section, we apply our approach to different communication methods and conduct experiments on various benchmarks. In specific, we aim to answer the following questions based on the experimental results in this section: 1) Can MA3C facilitate the robustness of multi-agent communication, and does each part in our approach make effect (Sec.~\ref{exp:main_performance})? 2) What kind of diverse attacker population has been obtained by MA3C (Sec.~\ref{exp:behavior_analysis})?  3) How does MA3C perform in the face of communication attacks with unseen perturbation ranges (Sec.~\ref{exp:transfer})?  Besides, we offer descriptions about the benchmarks in Sec.~\ref{exp:env} and do parameter sensitivity studies in Sec.~\ref{exp:parameter_study}.


Specifically, in our experiments, we apply our approach to three different communication algorithms: Full-Comm~\cite{guan2022efficient}, NDQ~\cite{ndq}, and TarMAC~\cite{tarmac}, which are of different features. Full-Comm is a popular communication paradigm, where each agent directly broadcasts their individual observations and updates the communication networks with an end-to-end scheme to minimize the Q-value estimation loss, showing competitive communication ability in multiple \correct{sceneriaos}{scenarios}~\cite{ijcai2022p82, guan2022efficient}. NDQ aims to generate meaningful messages and does message minimization to achieve nearly decomposable Q-value functions. TarMAC applies an attention mechanism in the receiving end to help agents focus on the specific part of the received messages. The details about NDQ and TarMAC are shown in App.~\ref{appx:baselines}.
\subsection{Environments}\label{exp:env}

\begin{figure*}[t]
    \centering
    \includegraphics[width=0.95\linewidth]{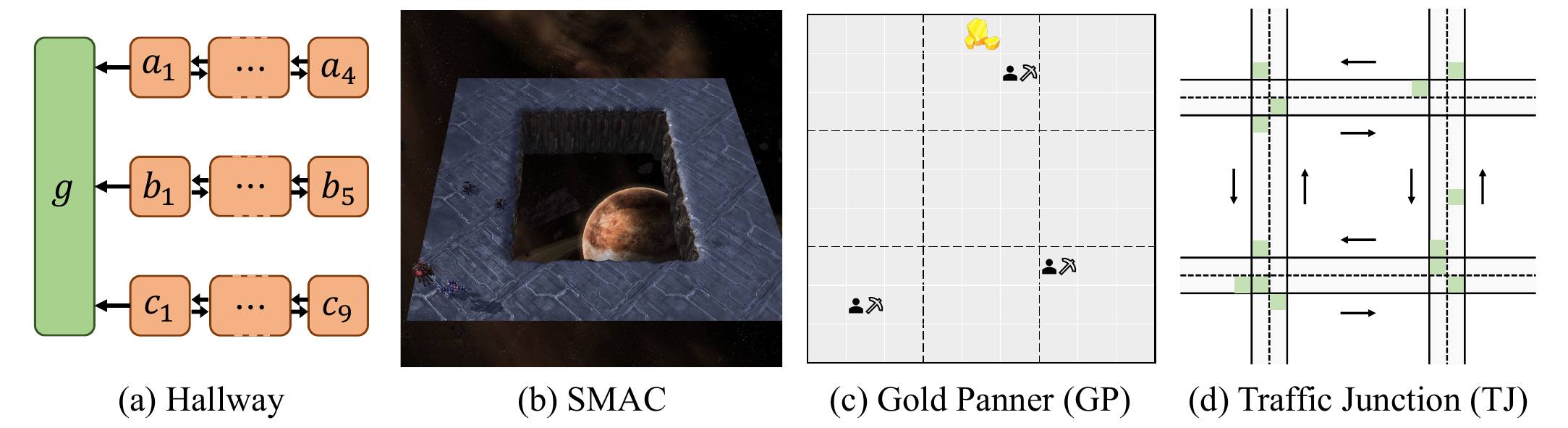}
    \caption{Experimental Environments used in this paper.}
    \label{fig:env}
\end{figure*}

In general, we select four multi-agent environments (see Fig.~\ref{fig:env}), respectively Hallway~\cite{ndq}, StarCraft Multi-Agent Challenge (SMAC)~\cite{smac}, a task environment we designed for multi-agent communication named Gold Panner (GP), and Traffic Junction (TJ)~\cite{tarmac}.
\paragraph{Hallway} Hallway is a multi-agent task with partial observability, where multiple agents are randomly spawned at different locations and required to reach the target simultaneously. In experiments, we design two instances of the Hallway task, where the first instance (Hallway-6x6) has two hallways with a length of 6, and the second instance (Hallway-4x5x9) has three hallways that have lengths of 4, 5 and 9, respectively. 
\paragraph{StarCraft Multi-Agent Challenge (SMAC)} StarCraft Multi-Agent Challenge (SMAC) is a popular benchmark for multi-agent cooperation, where there \correct{exist two camps of agent units in the scenario}{are agent units from two camps}, and the goal of the multi-agent algorithm is to control one of the camps to defeat the other. In particular, we select two maps named 1o\_2r\_vs\_4r and 1o\_10b\_vs\_1r from SMAC. In 1o\_2r\_vs\_4r, an Overseer is spawned around four enemy Reapers, and two ally Roaches are expected to reach the enemies and defeat them. Alike, in 1o\_10b\_vs\_1r, one Overseer detects a Roach, and 10 Banelings are required to reach and kill the enemies.
\paragraph{Gold Panner (GP)} To further validate our approach's effectiveness, we also design a task named Gold Panner (GP), which is a grid world with partial observability. This task divides the whole map into several parts, and the agents are randomly spawned at different regions. There exist one \correct{gold grid}{grid containing gold} that is initialized within sight of one agent, and agents are expected to load the gold at the same time. The core idea is that the agent nearby the gold ought to communicate the messages about the gold's location to the other agents to help all agents \correct{merge to}{gather near} the gold and load the gold together. We design two instances, which are respectively GP-4r and GP-9r. In GP-4r, there are 4 (2x2) regions, of which each region is a field with size of [3, 3], and three agents existing in the map, while, in GP-9r, there exist 9 (3x3) regions and 3 agents.
\paragraph{Traffic Junction (TJ)} Traffic Junction (TJ) is a familiar environment for testing the communication performance of multi-agent systems. In the task of Traffic Junction, multiple vehicles move on the two-way roads with several junctions and consistently follow a fixed route. We test on the medium version map of Traffic Junction, where the road dimension is 14, and there exist two junctions on each road when applying our approach to the communication method TarMAC.

\subsection{Robustness Comparison}\label{exp:main_performance}
\newcolumntype{C}[1]{>{\PreserveBackslash\centering}p{#1}}
\begin{table*}[t!]
\centering
\caption{Performance comparison under different attack modes.}\label{tab:main_exp}

\resizebox{\textwidth}{!}{
\begin{tabular}{C{2.4cm}C{2.4cm}C{2.4cm}C{2.4cm}C{2.4cm}C{2.4cm}C{2.4cm}C{2.4cm}}
\toprule
                     &                              & Hallway-6x6             & Hallway-4x5x9          & SMAC-1o\_2r\_vs\_4r    & SMAC-1o\_10b\_vs\_1r   & GP-4r                    &     GP-9r \\
\midrule
\multirow{5}{*}{Normal} & MA3C                      & 0.94$\pm$0.05           & 0.97$\pm$0.05          & 0.86$\pm$0.02          & 0.62$\pm$0.01          &   0.87$\pm$0.02          &     0.82$\pm$0.01          \\
                     & Vanilla                      & \textbf{1.00}$\pm$0.00  & \textbf{1.00}$\pm$0.00 & 0.81$\pm$0.06          & \textbf{0.63}$\pm$0.04 &   \textbf{0.88}$\pm$0.03 &     0.82$\pm$0.02          \\
                     & Noise Adv.                   & \textbf{1.00}$\pm$0.00  & 0.99$\pm$0.01          & \textbf{0.88}$\pm$0.04 & 0.6$\pm$0.05           &   \textbf{0.88}$\pm$0.03 &     \textbf{0.85}$\pm$0.02 \\
                     & MA3C w/o div.                & 0.98$\pm$0.02           & 0.66$\pm$0.46          & 0.86$\pm$0.02          & 0.62$\pm$0.03          &   0.86$\pm$0.09          &     0.81$\pm$0.03          \\
                     & Instance Adv.                & 0.52$\pm$0.48           & 0.67$\pm$0.47          & 0.84$\pm$0.02          & 0.57$\pm$0.04          &   0.86$\pm$0.03          &     0.82$\pm$0.03          \\
                     & \add{AME}                    & \add{\textbf{1.00}$\pm$0.00} & \add{0.98$\pm$0.02} & \add{0.81$\pm$0.05}  & \add{0.60$\pm$0.01}    &   \add{0.23$\pm$0.37}    &     \add{0.00$\pm$0.00}    \\
\cline{1-8}
\multirow{5}{*}{Random Noise} & MA3C & 0.91$\pm$0.07           & 0.79$\pm$0.18                         & \textbf{0.87}$\pm$0.01 & \textbf{0.67}$\pm$0.03 &   0.88$\pm$0.01          &     0.80$\pm$0.07          \\
                     & Vanilla                      & 0.58$\pm$0.03           & 0.53$\pm$0.06          & 0.73$\pm$0.07          & 0.60$\pm$0.02          &   0.86$\pm$0.03          &     0.79$\pm$0.02          \\
                     & Noise Adv.                   & \textbf{0.97}$\pm$0.02  & \textbf{1.00}$\pm$0.00 & 0.82$\pm$0.02          & 0.56$\pm$0.02          &   0.88$\pm$0.01          &     \textbf{0.82}$\pm$0.01 \\
                     & MA3C w/o div.                & 0.68$\pm$0.07           & 0.68$\pm$0.29          & 0.73$\pm$0.07          & 0.53$\pm$0.01          &   0.82$\pm$0.06          &     0.80$\pm$0.07          \\
                     & Instance Adv.                & 0.56$\pm$0.34           & 0.67$\pm$0.47          & 0.79$\pm$0.07          & 0.60$\pm$0.08          &   \textbf{0.90}$\pm$0.03 &     0.81$\pm$0.02          \\
                     & \add{AME}                    & \add{0.61$\pm$0.06}     & \add{0.79$\pm$0.03}    & \add{0.71$\pm$0.13}    & \add{0.59$\pm$0.08}    &   \add{0.22$\pm$0.37}    &     \add{0.00$\pm$0.00}    \\
\cline{1-8}
\multirow{5}{*}{Aggressive Attackers} 
                     & MA3C                         & \textbf{0.91}$\pm$0.22  & \textbf{0.98}$\pm$0.01 & \textbf{0.67}$\pm$0.03 & \textbf{0.62}$\pm$0.03 &  \textbf{0.81}$\pm$0.02  &     \textbf{0.76}$\pm$0.03 \\
                     & Vanilla                      & 0.09$\pm$0.19           & 0.00$\pm$0.00          & 0.26$\pm$0.12          & 0.57$\pm$0.03          &     0.38$\pm$0.02        &     0.30$\pm$0.05          \\
                     & Noise Adv.                   & 0.61$\pm$0.37           & 0.13$\pm$0.14          & 0.51$\pm$0.02          & 0.54$\pm$0.03          &     0.41$\pm$0.13        &     0.48$\pm$0.11          \\
                     & MA3C w/o div.                & 0.57$\pm$0.39           & 0.96$\pm$0.03          & 0.54$\pm$0.05          & 0.61$\pm$0.02          &     0.68$\pm$0.06        &     0.71$\pm$0.01          \\
                     & Instance Adv.                & 0.63$\pm$0.42           & 0.88$\pm$0.14          & 0.28$\pm$0.01          & 0.61$\pm$0.04          &  \textbf{0.81}$\pm$0.02  &     \textbf{0.76}$\pm$0.03 \\
                     & \add{AME}                    & \add{0.13$\pm$0.03}     & \add{0.00$\pm$0.00}    & \add{0.39$\pm$0.05}    & \add{0.59$\pm$0.07}    &   \add{0.10$\pm$0.16}    &     \add{0.00$\pm$0.00}    \\
\bottomrule
\end{tabular}
}
\end{table*}

To testify whether our approach can facilitate robust communication when applied to different communication algorithms and scenarios, we apply our approach to three communication methods and select four tasks requiring agent communication. Specifically, we employ Full-Comm in Hallway, SMAC, and GP tasks, and NDQ and TarMAC in SMAC and TJ, respectively. The test results are listed in Tab.~\ref{tab:main_exp}, and more details about the experiments are provided in App.~\ref{appx:experimentaldetail}.

For the compared algorithms, we design three straightforward baselines, including Vanilla, Noise Adv. and Instance Adv.
respectively. Vanilla does not apply any adversarial training technique and learns the ego system policy in scenarios without communication attacks. Noise Adv. applies adversarial training with random noise attacks. While Instance Adv. builds one single communication attacker, training the communication system and the attacker alternatively. 
\add{Actually, Instance Adv. can be seen as an ablation that does not use population, thus used to verify the influence of attacker population.}
One special note is that to alleviate the overfitting problem of adversarial training with one single attacker and thus construct a stronger baseline, we enhance Instance Adv. by maintaining a pool of historical attacker models and doing adversarial training against the whole pool.
\correct{Moreover, to further justify the role of our diversity mechanism for the attacker population, we further design an ablation named MA3C w/o div. that \correct{only dumps}{simply does not employ} the diversity mechanism compared with MA3C.
}{Besides, to make our work more solid, we additionally compare our approach with two existing methods for robustness. The first one is the variant of RAP in our experiment setting, which also adopts the adversary population for training robust policies. Its main difference from our approach is that it does not explicitly optimize the diversity of the population. Thus, we denote this baseline as MA3C w/o div., which can be used to verify the effectiveness of our diversity mechanism. The other compared method is a recently proposed work called AME~\cite{sun2022certifiably}, which builds a message-ensemble policy by aggregating the decision results of utilizing different ablated message subsets.}

For each method, we employ three different test modes: 1) \textbf{Normal} means no communication attacks, which is to show the communication performance of the multi-agent system in clean scenarios; 2) \textbf{Random Noise} tests the communication robustness under random noise attacks; 3) \textbf{Aggressive Attackers} additionally trains a set of unseen communication attackers and utilizes them to do robustness test.

As we can see from Tab.~\ref{tab:main_exp}, our approach MA3C exhibits comparable or more \correct{incredible}{better} performance \correct{to}{than} other baselines when applied to Full-Comm. Concretely, the Vanilla baseline, trained in a natural manner, performs excellently in the Normal setting where no noise exists but suffers from performance drop when tested in a noisy communication environment. It even fails when encountering aggressive attackers (e.g., it obtains zero success rate on Hallway-4x5x6). This phenomenon reveals the vulnerability of communication-based policy and calls for algorithm design to enhance communication robustness. As for Noise Adv., it works well when tested with random noise attacks since this corresponds to its training situation. However, it struggles when encountered with unseen aggressive attackers, e.g., a performance drop of $0.87$ is found in the task of Hallway-4x5x9. \correct{We hypothesize the reason behind it is that the random generated noise attacks can hardly cover some specific attack patterns, and the obtained policy fails when facing those-like communication attacks.}{We hypothesize that the reason for this is that the randomly generated noise attacks can hardly cover some specific attack patterns, and the obtained policy fails in the face of such communication attacks.} Furthermore, for the Instance Adv., though we strengthen it by maintaining a pool of historical attacker models, it still suffers from coordination degradation in the Normal situation in different environments, which shows that adversarial training may \correct{injury}{damage} the communication ability, and has been discussed in some other RL domains~\cite{vinitsky2020robust}. The performance advantage of MA3C and MA3C w/o div. over Instance Adv. demonstrates the effectiveness of population.
 \begin{figure*}[t!]
    \centering
    \includegraphics[width=\linewidth]{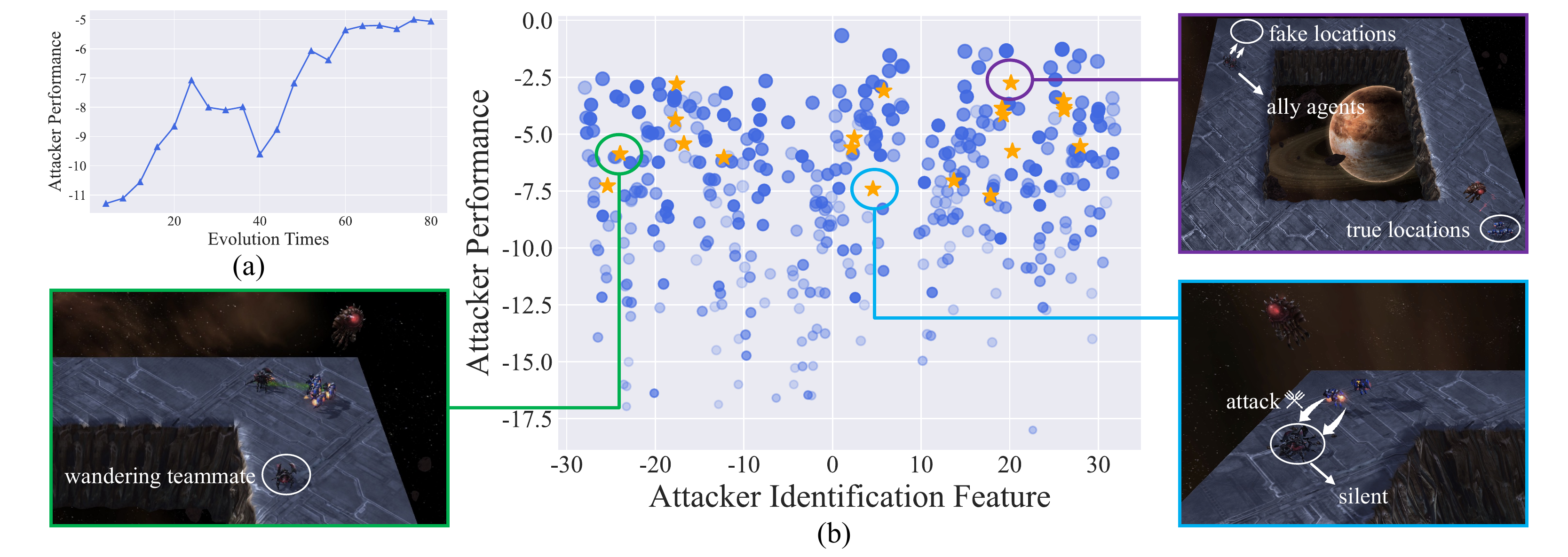}
    \caption{Population visualization. In specific, each scatter corresponds to an attacker instance, and we use the color depth to represent the training stage of the attackers, i.e., the lighter the color, the earlier the attacker model. The horizontal coordinate indicates the identification feature after dimension reduction, and the vertical coordinate indicates the attack performance of the attacker model.}
    \label{fig:pop_visualization}
\end{figure*}

 \begin{figure}[t!]
    \centering
    \includegraphics[width=0.95\linewidth]{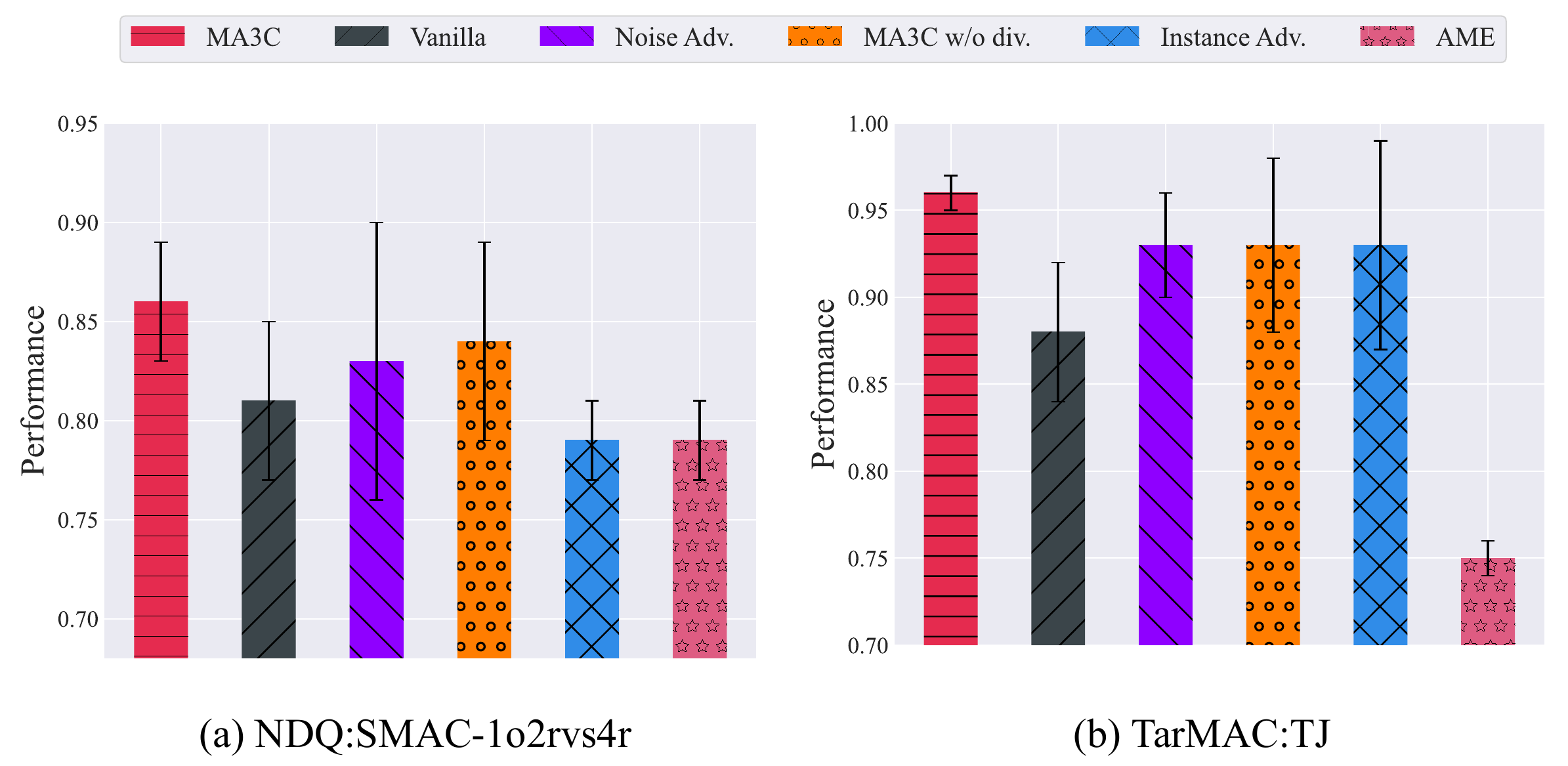}
    \caption{Robustness comparison when employing NDQ on SMAC-1o\_2r\_vs\_4r and TarMAC on TJ, respectively.}
    \label{fig:al_res}
\end{figure}


Besides, the comparison to the baseline MA3C w/o div.\add{, which conducts a similar approach to RAP in our setting,} proves that our diversity mechanism has good gains in the robustness of obtained ego system policy. The essential reason behind it is that it can avoid homogeneity of instances in the population and ensure that the communication attackers encountered during the training phase can cover more attack patterns. 
\add{Again, when compared to the baseline AME, our approach still exhibits good performance advantages. By aggregating the decision results of utilizing multiple ablated message subsets, AME can somewhat alleviate the influence of communication attacks and shows relatively good performance when tested with random noise on the environments of Hallway and SMAC. However, when we apply aggressive attackers to test its robustness, its performance faces a significant drop, which shows the limitation of this kind of approach. Besides, we find that AME performs terribly on the environment of GP. We hypothesize the reason is that some specific channels are vital for the agents to complete this task, so the practice of utilizing ablated message subsets may miss these critical messages. Considering that the AME approach is more suited to the setting where a portion of agents are attacked, we additionally test MA3C and AME in this setting, and the results are reported in App.~\ref{appx:extra_experiment}. It can be seen that, even in this test setup, MA3C still shows better communication robustness, which further justifies the superiority of our approach.} 

Furthermore, since our approach and other baselines are all agnostic to specific communication methods, we also implement them on other typical communication methods such as NDQ and TarMAC. As can be seen from Fig.~\ref{fig:al_res}, the superiority over other baselines demonstrates the generality of MA3C.


\subsection{Attacking Behavior Analysis}\label{exp:behavior_analysis}



To reveal what kinds of attackers have been obtained by our approach, we conduct visualization analysis 
to check whether our attacker population optimization method can help obtain a population with diverse and qualified attackers in the task of SMAC-1o\_2r\_vs\_4r. Specifically, we pre-train a multi-agent communication system and apply our population mechanism to train an attacker population to conquer the communication system. Each attacker instance in the population has an identification vector, as mentioned before. We take out attacker instances from the population at various stages throughout the training process, downscale their feature vectors to one dimension, and visualize them in Fig.~\ref{fig:pop_visualization}. We expect the attackers to cover more regions along the horizontal coordinate, which means diversity, and \correct{are located}{to be located} as high as possible for the vertical coordinate, indicating good attack ability.

From the results shown in Fig.~\ref{fig:pop_visualization}, we can see that scatters on the top tend to be darker, implying that the population optimization process can help obtain stronger attackers. We also compute the whole attack performance, which equals to the average attack performance of all attackers in the population, and plot the variant curve in Fig.~\ref{fig:pop_visualization}(a), from which we can see the upward trend of the population's attack ability. Besides, we mark the final population in the last iteration as yellow stars, and we can see that the final $20$ attackers are diverse along the attacker identification feature axis. To further check what attack patterns have been learned for the attackers in the population, we render the trajectories for specific attacker instances and find: (1) in the picture at the bottom left, one Roach wanders around, leaving its teammate alone to battle with the enemies; (2) in the picture at the top right, the messages of the Overseer are attacked, and fake enemy location information is transmitted to the two teammates, leading to the two ally Roaches moving towards the fake locations; (3) in the picture at the bottom right, attacker achieves effective attacks by tricking the allies into remaining silent during battling through message perturbations. 
\begin{figure}[t!]
\subfigure[SMAC-1o\_2r\_vs\_4r]{
\centering
\includegraphics[width=0.45\linewidth]{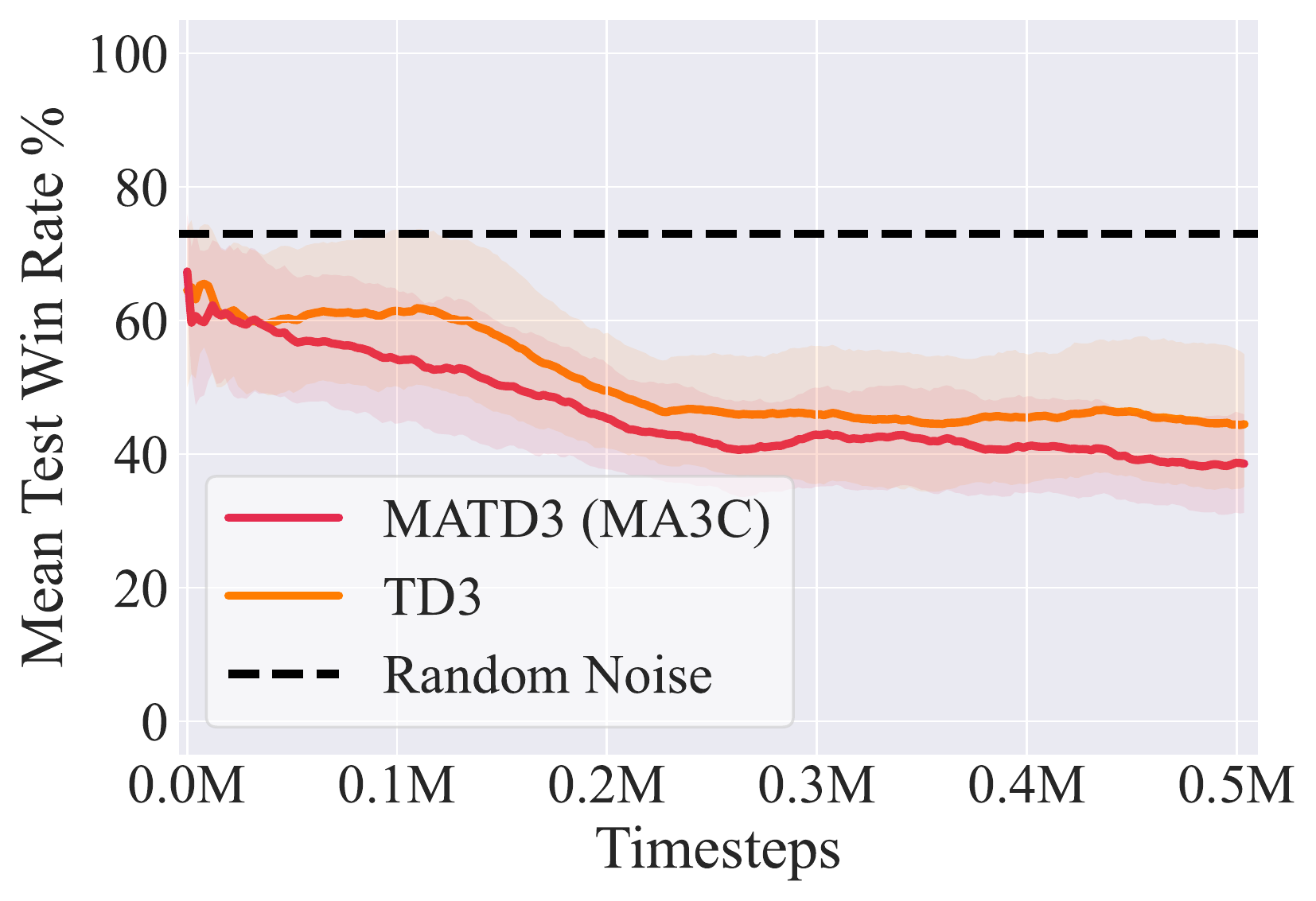}
}
\subfigure[GP-4r]{
\includegraphics[width=0.45\linewidth]{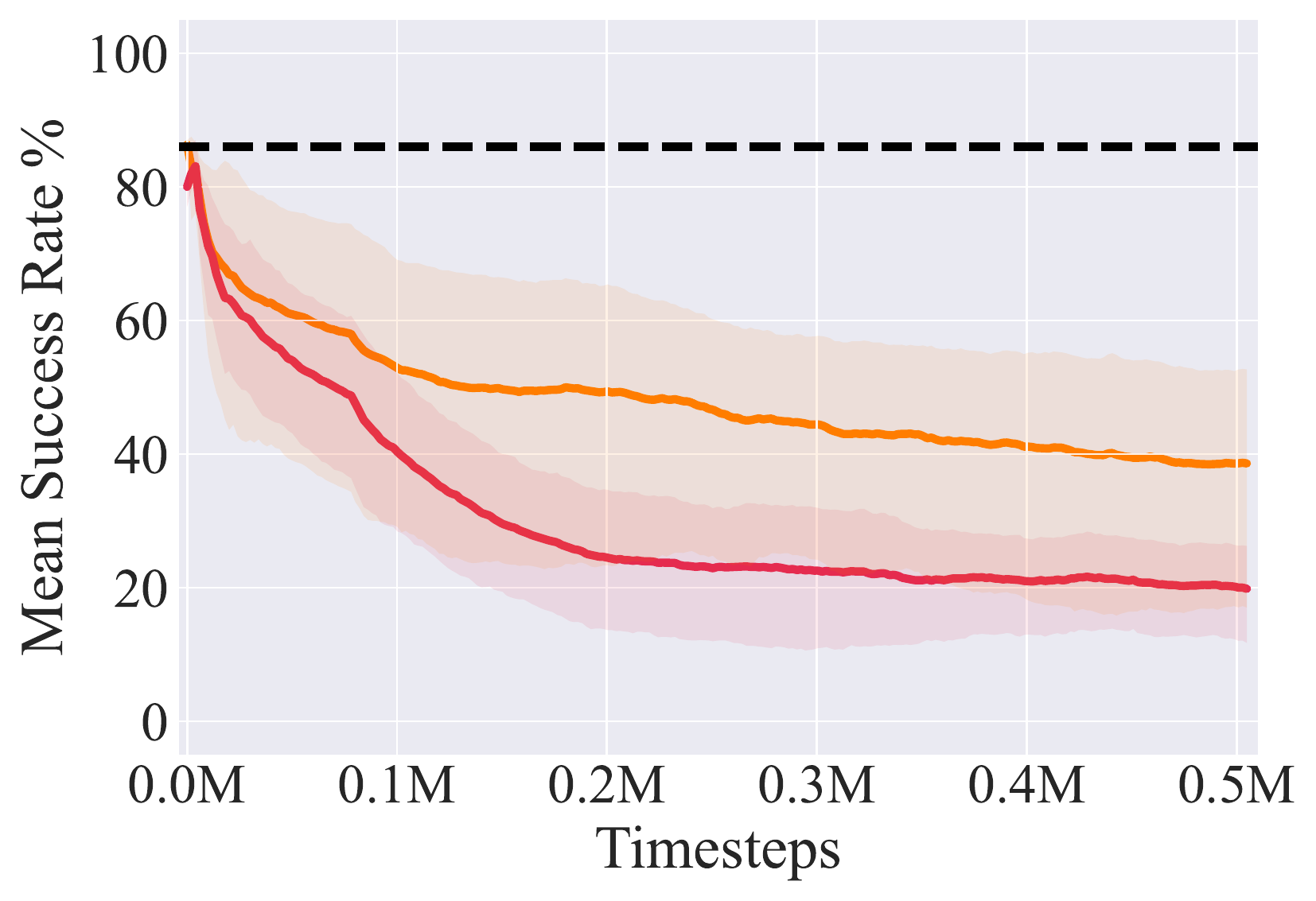}
}
\caption{Comparison of the attack ability of different methods.}
\label{fig:aa}
\end{figure}

After that, we conduct experiment to investigate whether our approach can help obtain effective attackers. The concern is that the adversarial training will be meaningless if the obtained attackers can not generate practical communication attacks. Thus to confirm the validity of the attacker learning, we independently train multiple attackers with our approach in tasks of SMAC-1o\_2r\_vs\_4r and GP-4r. In both tasks, we restrict the learning process to be within $500K$ samples, and the vertical coordinate indicates the test performance of the attacked ego system. To further justify our practice of learning the attacker via the MARL method, we additionally compare with a baseline that learns the attacker with the TD3~\cite{DBLP:conf/icml/FujimotoHM18} algorithm, which means treating it as a single-agent learning problem. A baseline named Random Noise is also added to show the performance under random noise attacks. From the learning curves in Fig.~\ref{fig:aa}, we can see that the TD3 baseline and our approach can \correct{coverage}{converge} to much lower performance compared with Random Noise, which means that they have found the vulnerability of the communication system and learned effective attack patterns. Also, our approach exhibits better attack performance than the TD3 baseline, which shows the effectiveness of modeling the attacker learning as a multi-agent learning problem, and the virtual agents actually learn to cooperate to attack the communication system, obtaining better attack performance.


\subsection{Policy Transfer}\label{exp:transfer}
\begin{figure}[t!]
\subfigure[SMAC-1o\_2r\_vs\_4r]{
\centering
\includegraphics[width=0.45\linewidth]{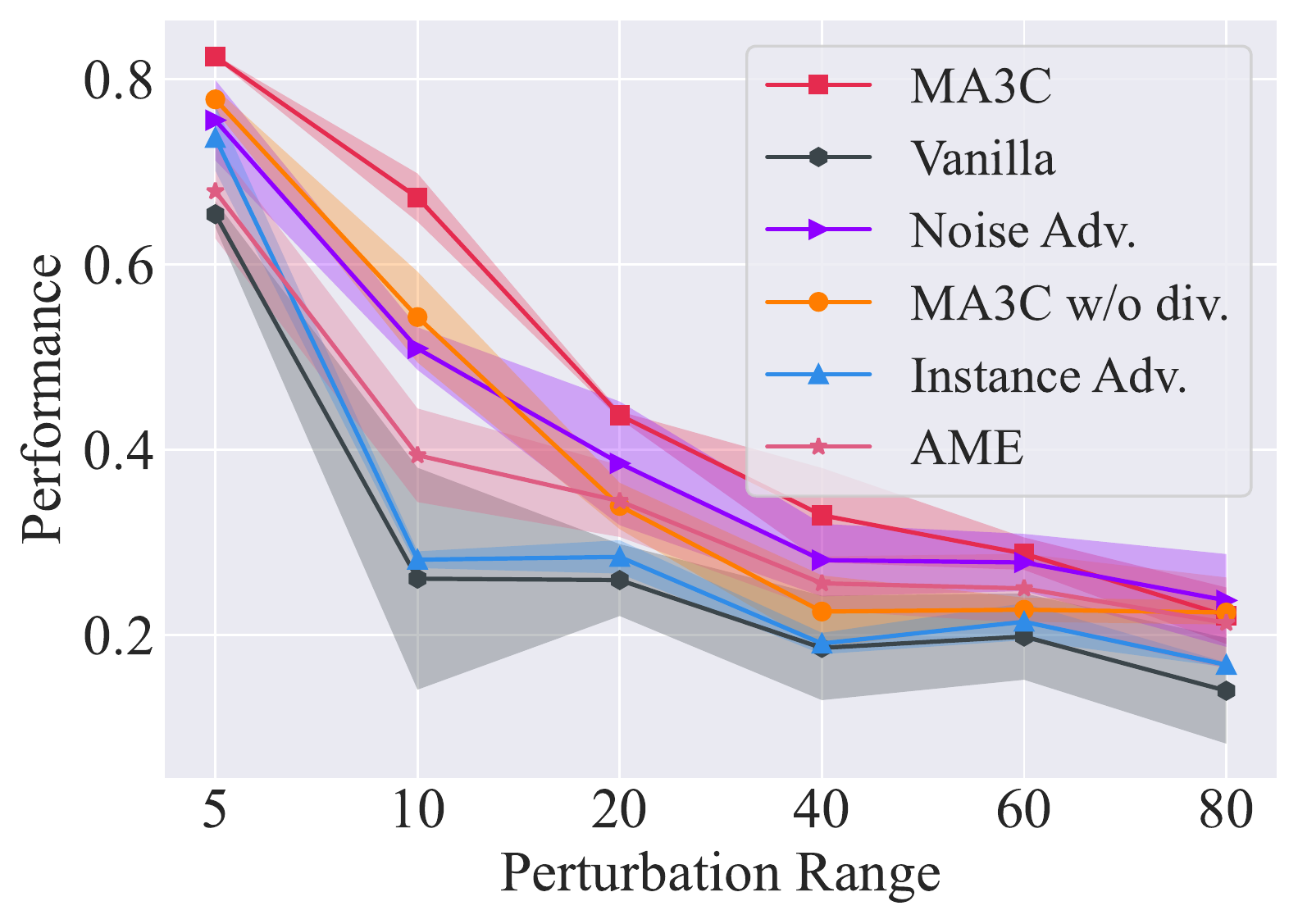}
}
\subfigure[GP-4r]{
\centering
\includegraphics[width=0.45\linewidth]{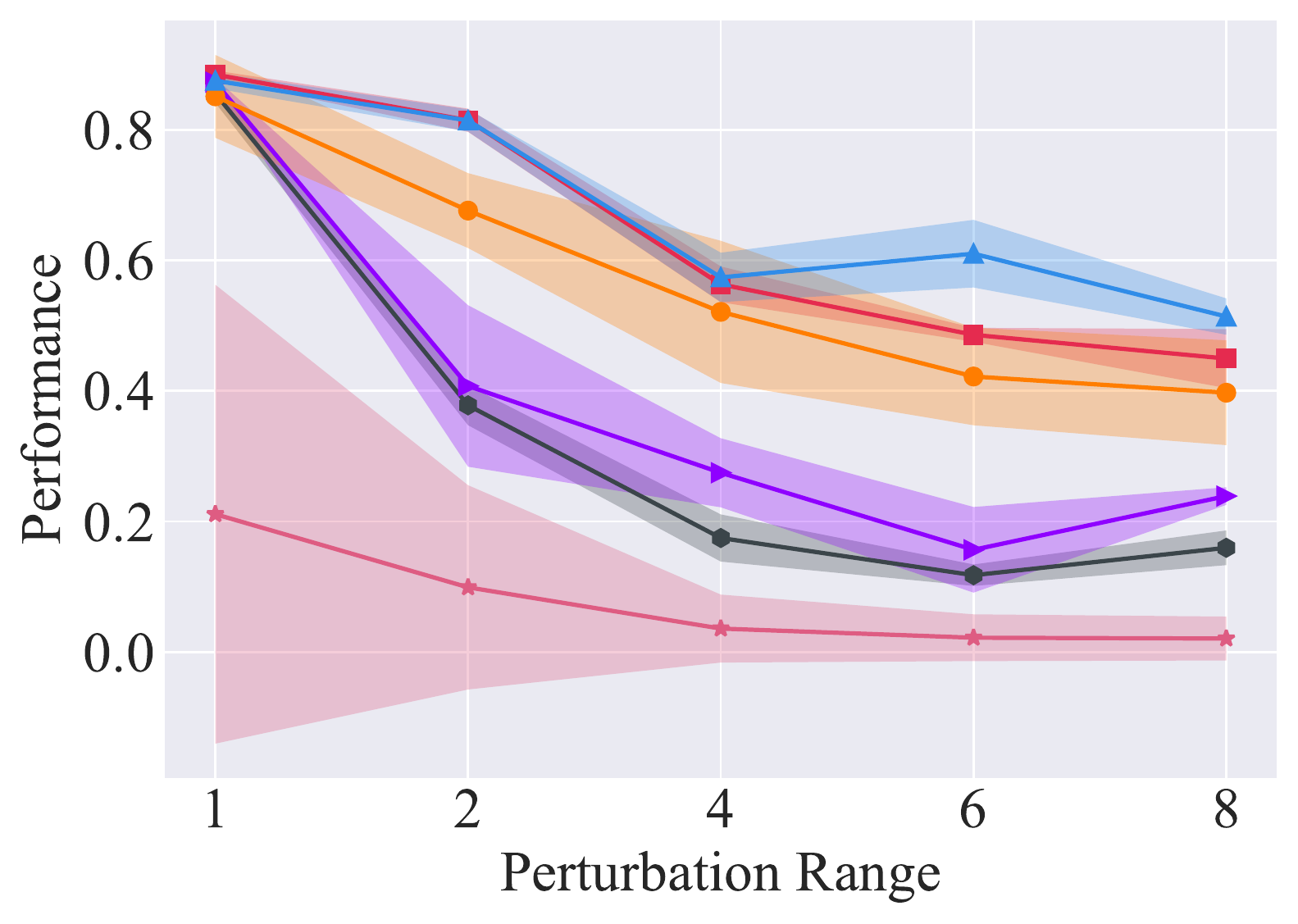}
}
\caption{Generalization test to different perturbation ranges.}
\label{fig:generalization}
\end{figure}
In fact, we suppose that the allowed perturbations are always in a restricted set $B(m)$, and all the experiments above assume that the test perturbation range is the same as that utilized in the adversarial training, of which the details are introduced in App.~\ref{appx:experimentaldetail}. However, in many practical scenarios, we can not suppose the real communication attacks encountered in the execution phase are always within the perturbation range designed in the training phase. Thus, we wonder how our approach performs when generalizing to attacks with different perturbation ranges.

\begin{figure}[t!]
\subfigure[Hallway-4x5x9]{
\includegraphics[width=0.45\linewidth]{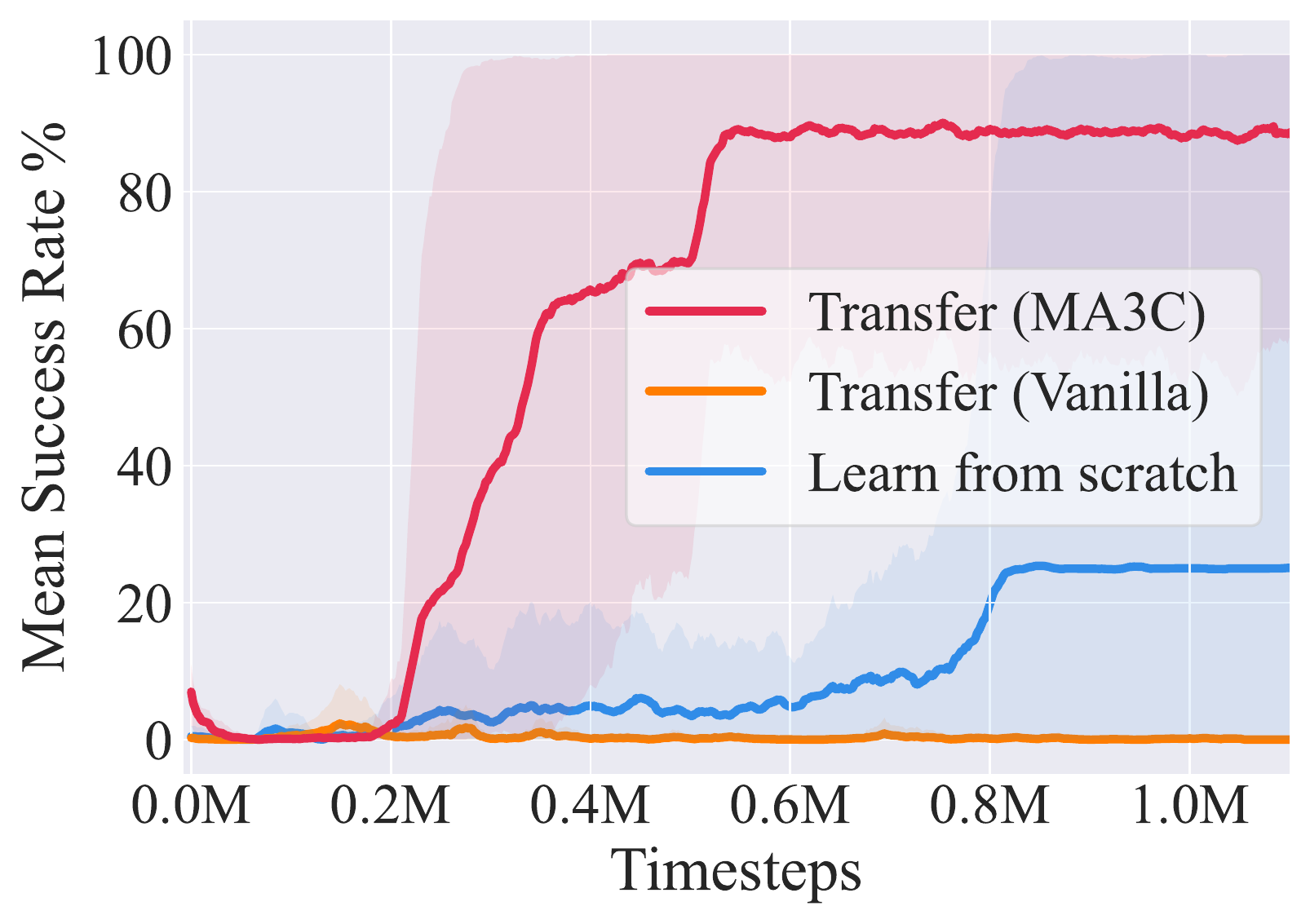}
}
\subfigure[GP-4r]{
\includegraphics[width=0.45\linewidth]{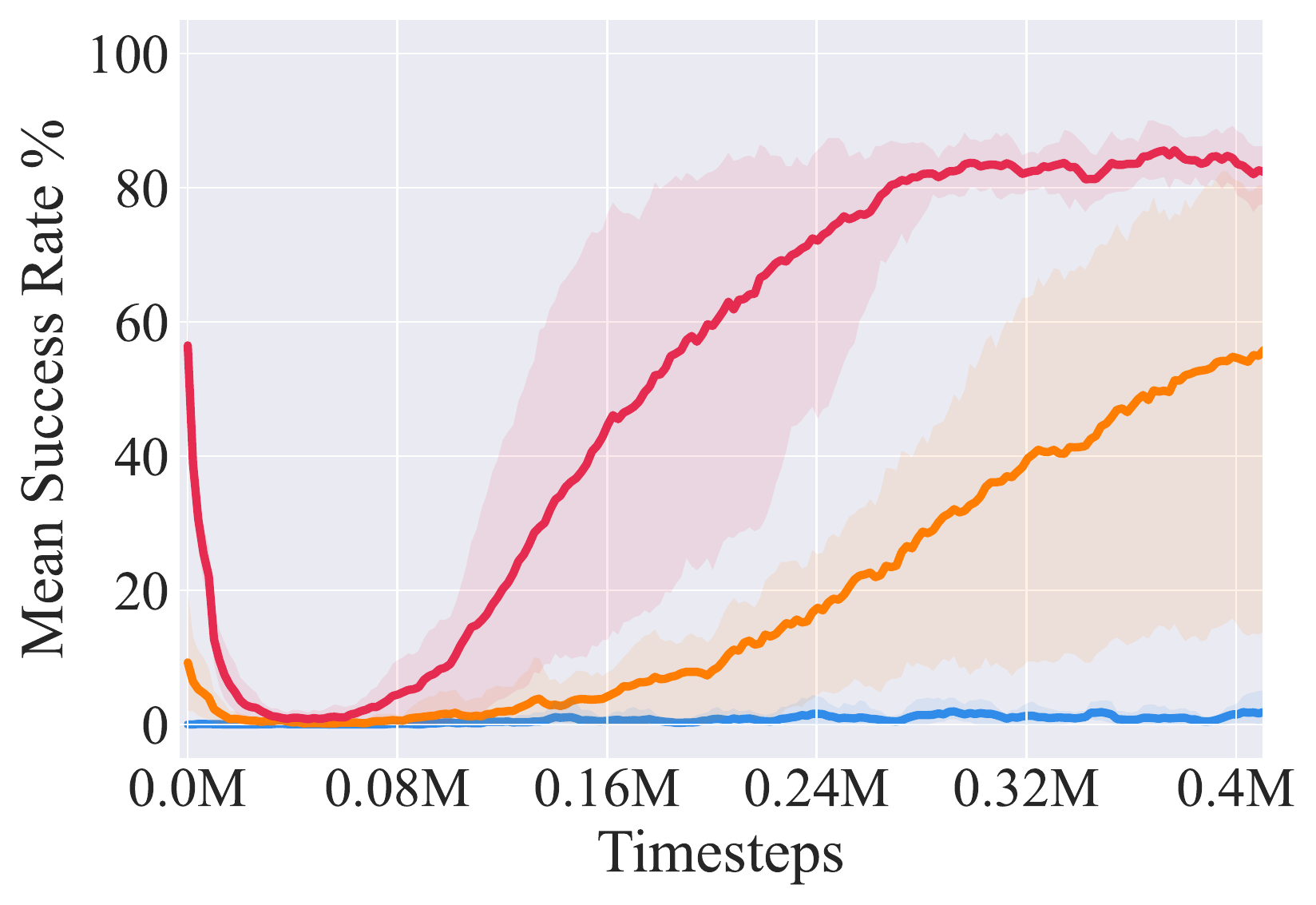}
}
\caption{Transfer to larger perturbation range.}
\label{fig:transfer}
\end{figure}

Firstly, we collect multiple aggressive attackedrs using the same method in Sec.~\ref{exp:main_performance}, but with different perturbation ranges when testing, which can be seen as a direct zero-shot generalization test. We conduct the experiment in tasks of SMAC-1o\_2r\_vs\_4r and GP-4r, and the results are recorded as plots in Fig.~\ref{fig:generalization}. From the results, we can see that as the perturbation range increases, all algorithms face a consistent performance degradation trend. This phenomenon is expectable because when the perturbation range is larger, greater changes in the input values tend to have a greater impact on the network's output, thus causing larger harm to the communication performance. Besides, we find our approach \name~exhibits good generalization performance under different unseen perturbation ranges. For example, in the task of SMAC-1o\_2r\_vs\_4r, MA3C still obtains the highest win rate compared to baselines and ablations under different perturbation ranges. This demonstrates the generalization ability of our approach, which is of significance for the deployment in some unknown scenarios.
\begin{figure*}[t!]
\centering
\subfigure[Studies of Population Size]{
    \includegraphics[width=0.3\textwidth]{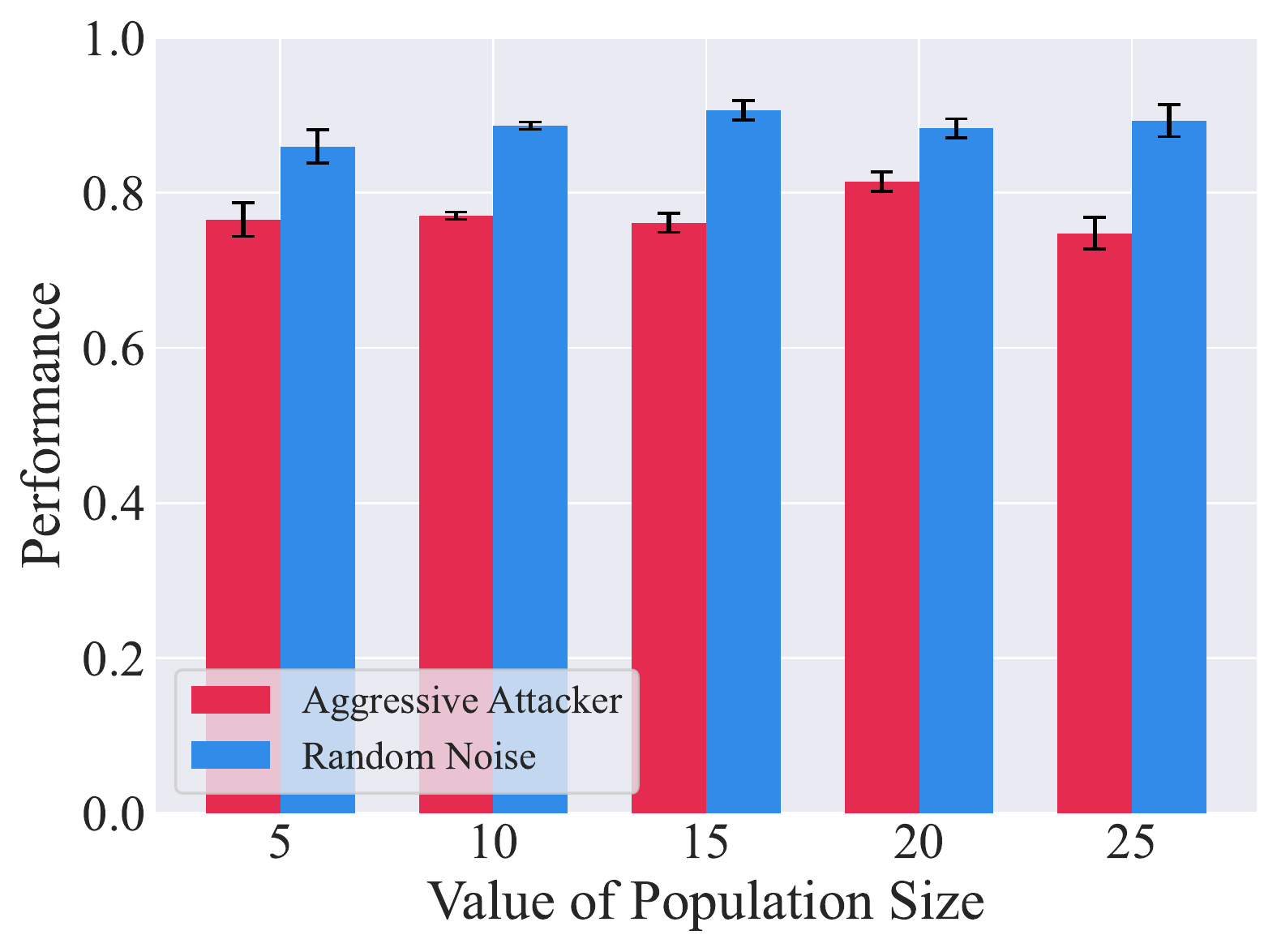}
   }
\subfigure[Studies of Reproduction Ratio]{
    \includegraphics[width=0.3\textwidth]{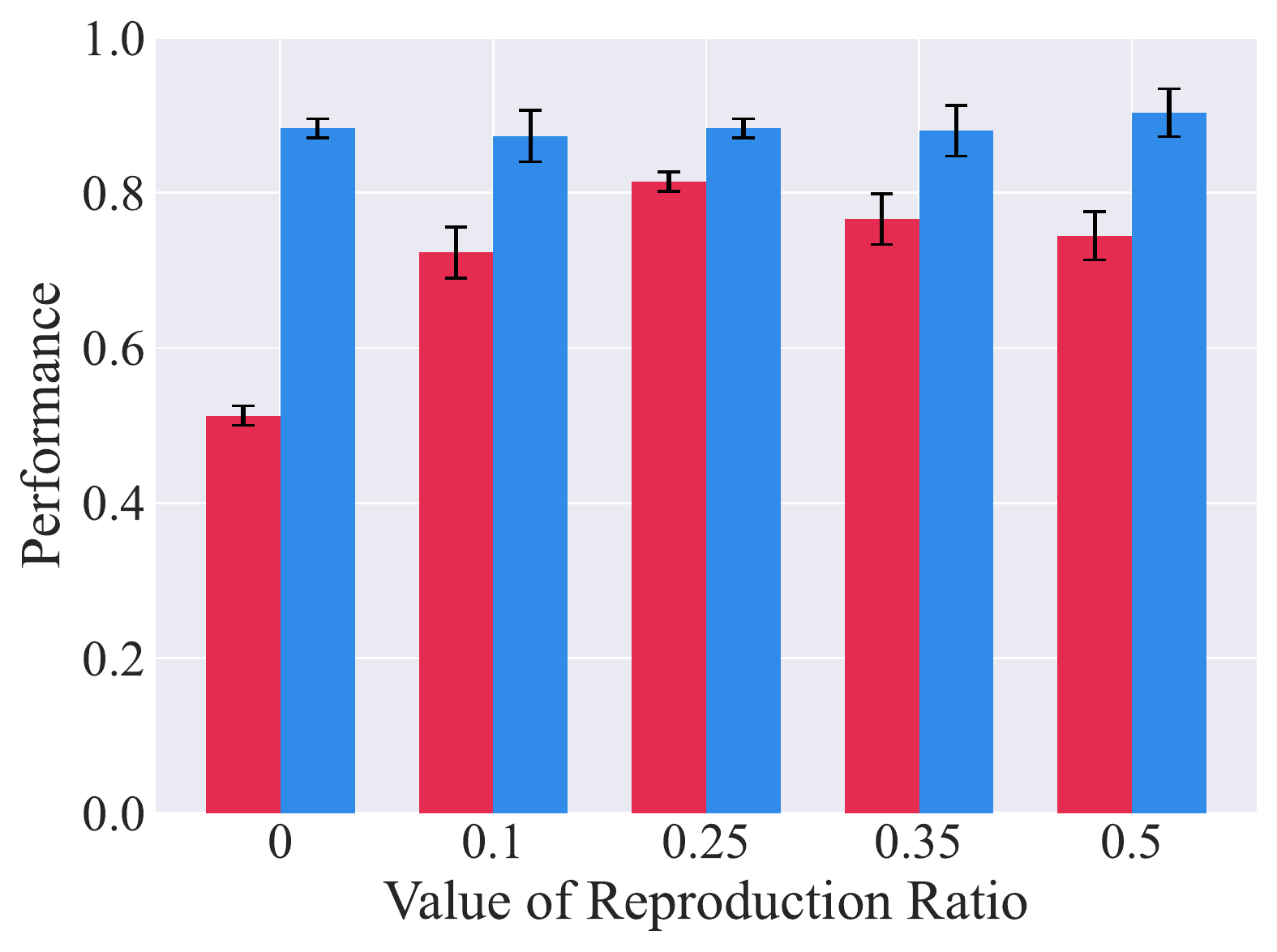}
   }
\subfigure[Studies of Distance Threshold]{
    \includegraphics[width=0.3\textwidth]{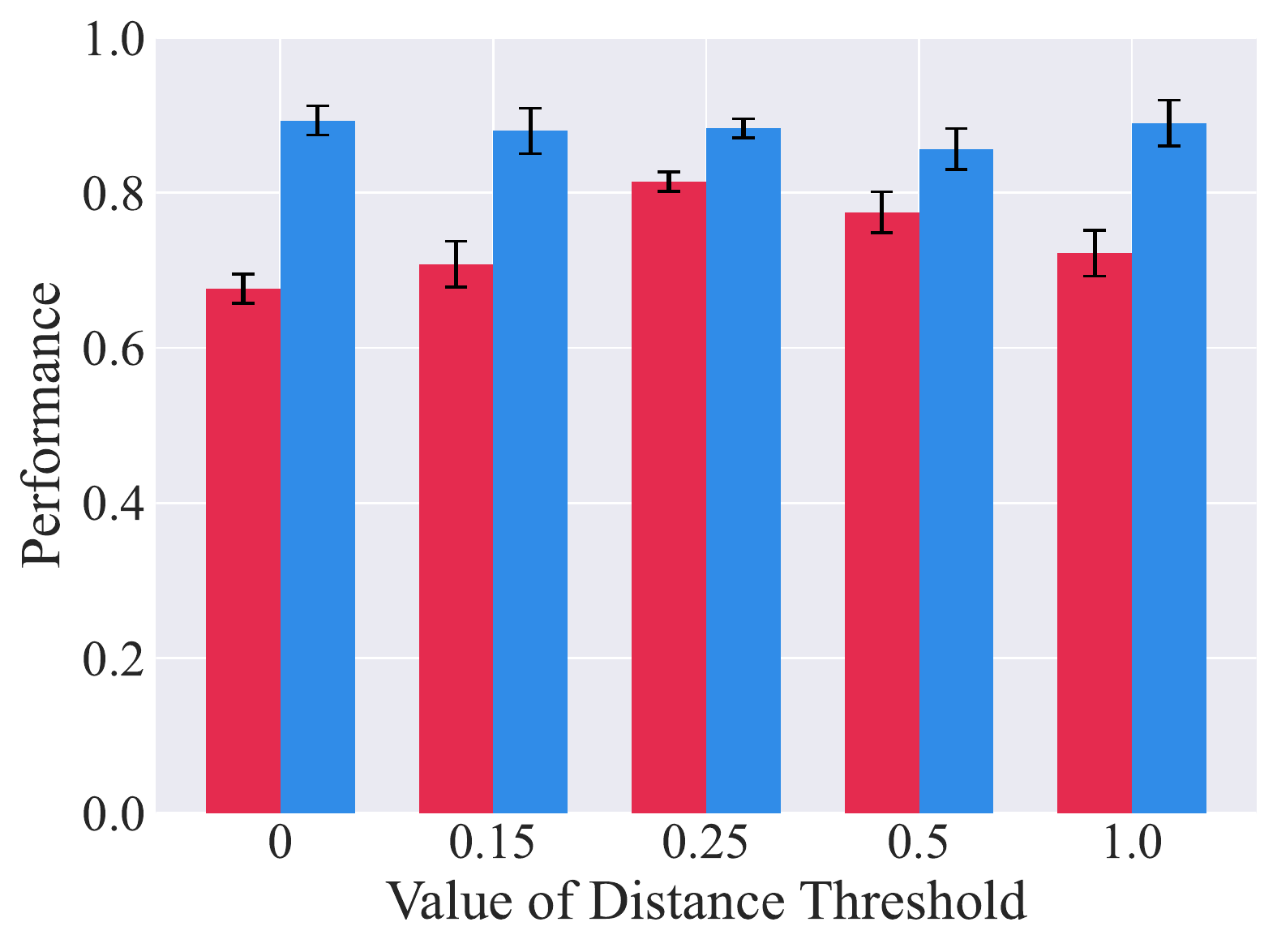}
    }
\caption{Test results of parameter sensitivity studies.}
\label{fig:param}
\end{figure*}

Furthermore, we also conduct experiments to look into whether our approach \name~can help achieve a quick adaptation to new environment. We here conduct transfer experiments to quite large perturbation ranges to test this property. Specifically, we select perturbation ranges of $5$ and $15$ in the tasks of Hallway-4x5x9 and GP-4r, respectively, compared to choices of $1.0$ and $2$ during adversarial training. To validate the effectiveness of the our approach, we compare with the baseline which learns from scratch in the noisy scenarios. Besides, to certify that the transfer gain is not from the extra pre-training, we add a baseline that only pre-trains the ego system policy in a clean scenario. The results in Fig.~\ref{fig:transfer} demonstrate that \name~is equipped with a good property for transferring to larger perturbation ranges, and we claim that it is of great value for some real-world applications. 

\subsection{\add{Trajectory Encoder Studies}}\label{exp:traj_encoder study}

\begin{figure}[t!]
    \centering
    \includegraphics[width=\linewidth]{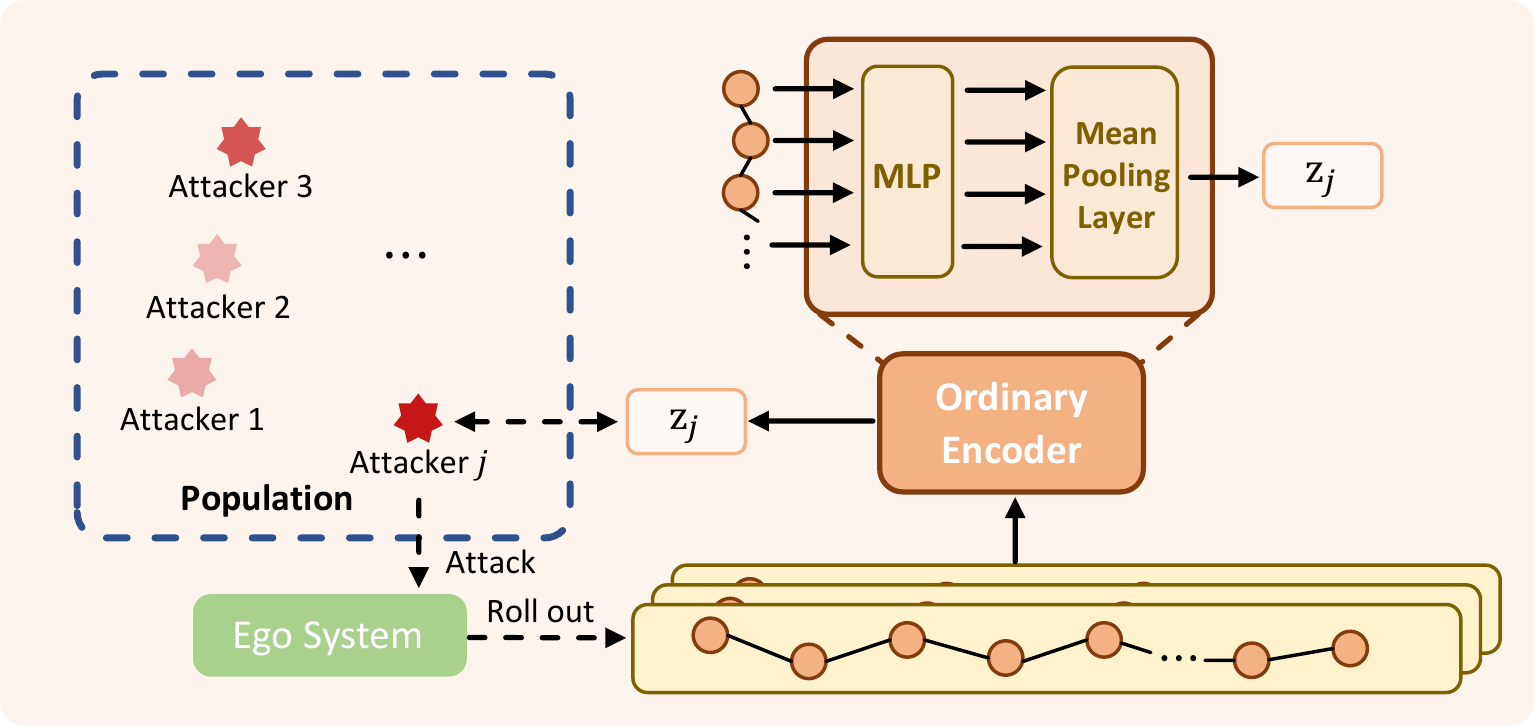}
    \caption{The architecture of the ordinary trajectory encoder. We feed the state information at each time step into a shared MLP network, and then perform mean pooling on the outputs to obtain the embedding vector of the trajectory.}
    \label{fig:dummy_traj_encoder}
\end{figure}

\add{
In our approach, we adopt the transformer architecture as the trajectory encoder that helps distinguish the attack patterns of different attacker instances. The superiority of the transformer architecture is that its attention mechanism can help capture the critical points in the trajectory, and its network expressiveness can help handle complex scenarios. To further demonstrate the advantages of the transformer architecture, we additionally consider a variant of \name~that uses Multi-Layer Perception (MLP) and Mean-Pooling technology to obtain the trajectory representation as shown in Fig.~\ref{fig:dummy_traj_encoder}. In this variant, the transformer architecture of the trajectory encoder is removed. We use this experiment to study how much influence the transformer architecture has.
}

\add{
We apply this variant to the GP environment, of which the results are shown in Tab.~\ref{tab:traj_enc}. Note that this variant differs from \name~only in the trajectory encoder network and all other training details are the same. 
In fact, we can see from the results that MA3C with ordinary trajectory encoder achieves comparable performance to MA3C when tested with random noise attacks. However, when tested with aggressive attackers, this variant suffers a greater drop in performance compared to MA3C, but still shows better robustness than MA3C w/o div. slightly. For example, when tested with aggressive attackers, MA3C, MA3C w/ordinary encoder, and MA3C w/o div. suffer performance drops~\footnote{\add{The performance drop rate here is calculated as the percentage of drop in performance when moving from the Normal mode to being attacked by Aggressive Attackers, i.e. $\frac{p(\text{Normal}) - p(\text{Aggressive Attackers})}{p(\text{Normal})}$, where $p(X)$ is the performance in the test mode of $X$.}} of 7\%, 19\%, and 21\%, respectively. We hypothesize that this is because the ordinary encoder fails to capture the differences between certain trajectories, thus interfering with MA3C's diversity mechanism. This results in the variant failing to cover some specific attack patterns and causing a greater performance drop.
}

\begin{table}[t!]
\centering
\caption{Test results for the trajectory encoder studies.}\label{tab:traj_enc}
\begin{tabular}{C{1cm}C{2.4cm}C{2cm}C{2cm}C{2cm}}
\toprule
                     &                              & MA3C             & \makecell{MA3C w/\\ordinary encoder}          & MA3C w/o div.   \\
\midrule
\multirow{3}{*}{GP-4r} & Normal                      & {0.87$\pm$0.02}           & {0.90$\pm$0.02}          & {0.86$\pm$0.09}     \\
                     & Random Noise                      & {0.88$\pm$0.01}           & {0.90$\pm$0.03}          & {0.82$\pm$0.06}     \\
                     & Aggressive Attackers                   & {0.81$\pm$0.02}           & {0.73$\pm$0.04}          & {0.68$\pm$0.06}     \\
\cline{1-5}
\multirow{3}{*}{GP-9r} & Normal                      & {0.82$\pm$0.01}           & {0.82$\pm$0.03}          & {0.81$\pm$0.03}     \\
                     & Random Noise                      & {0.80$\pm$0.07}           & {0.80$\pm$0.05}          & {0.80$\pm$0.07}     \\
                     & Aggressive Attackers                   & {0.76$\pm$0.03}           & {0.70$\pm$0.06}          & {0.71$\pm$0.01}    \\
\bottomrule
\end{tabular}
\end{table}

\subsection{Parameter Sensitivity Studies}\label{exp:parameter_study}

Finally, to investigate how the parameters introduced in our work influence the final robust performance of our approach, we selectively conduct parameter sensitivity studies in the task of GP-4r.  Specifically, we choose three core hyper-parameters: (1) Population size: the size of the population; (2) Reproduction Ratio: the ratio of attackers each time we select to do evolution; (3) Distance Threshold: a hyper-parameter to determine whether a new attacker instance is novel enough for the current population. We here report the results under Aggressive Attackers and Random Noise. Note that the default hyper-parameter selection in the experiments above is listed in App.~\ref{appx:implementationdetail}, where the default value for Population, Reproduction Ratio and Distance Threshold are respectively $20$, $0.25$ and $0.25$.

As we can see from the results in Fig.~\ref{fig:param}, overall the performance of the communication system under Aggressive Attackers is much lower than that under Random Noise, and the performance under Random Noise is not sensitive to the hyper-parameters, which confirms that Aggressive Attackers have higher attack ability than Random Noise. Among these three hyper-parameters, we find that the Population Size has the least impact on the algorithm performance, implying that a population with size of five attacker instances works well in this task. For Reproduction Ratio, an interesting phenomenon is that the approach works poorly when Reproduction Ratio is zero. This phenomenon is because when Reproduction Ratio is zero, the population is always a set of random initialized attacker instances, which can provide limited information for adversarial training. Besides, for Distance Threshold, we find a selection of value $0.25$ works best in this task. A too small Distance Threshold like zero ignores the process of diverse selection, and this makes the algorithm degenerate to the baseline of MA3C w/o div., while a too-large Distance Threshold causes our algorithm to pay less attention to the attack ability of the attacker instances.

\section{Final Remarks}
How to obtain a robust communication-based policy recently became an emergent for policy deployment. Instead of employing existing techniques to get a robust policy under some constraints, we take a further step for this issue by learning adaptable multi-agent auxiliary adversaries to promote robustness for communication-based policy. Sufficient experiments conducted on multiple cooperative benchmarks demonstrate the high robustness ability of our proposed method, other results also show its high generalization ability for various perturbation ranges, and the learned policy can transfer learned
robustness ability to new tasks after fine-tuning with a few samples. As we consider a limited perturbation set, how we can develop an autonomous paradigm like curriculum learning to find the communication ability boundary is an invaluable direction, and developing efficient and effective MARL communication methods under the open-environment scenarios~\cite{zhou2022open} is challenging but of great value in the future.

\section*{Acknowledgement}
This work is supported by the National Key Research and Development Program of China
(2020AAA0107200), the National Science Foundation of China (61921006, 61876119, 62276126), the Natural Science Foundation of Jiangsu (BK20221442), and the program B for Outstanding PhD candidate of Nanjing University. We thank Ziqian Zhang and Lihe Li for their useful suggestions.




\section{Appendixes}

\subsection{Introduction to The Selected Baselines} \label{appx:baselines}

\textbf{NDQ}~\cite{ndq} considers that many multi-agent tasks in the real world are not fully decomposable and then achieve nearly decomposable Q-functions via communication minimization. Specifically, it trains the communication-based policy by maximizing the mutual information between the agents’ action selection and the message sent to the corresponding teammate. It also minimizes the entropy of messages between agents to avoid distribution collapse. Each agent broadcasts messages to all other agents and uses the received message to augment the local policy. As NDQ minimizes the message entropy, it can extract the most useful part for decision-making and shows great performance on many tasks like SMAC.

\noindent\textbf{TarMAC}~\cite{tarmac}  is a widely used MARL communication approach, which applies an attention mechanism to extract the most valuable information from multiple received messages. Concretely, each agent generates signature and query vectors as the message, and the message is then broadcasted to all teammates. In the message-receiving phase, the attention weights of incoming messages are obtained by calculating the similarity between the query vector and the signature vector of each incoming message. Then a weighted sum over all incoming messages is performed to determine the message an agent. The extracted message is 
finally used to augment the local observation for decentralized execution. 


\subsection{Implementation Details} \label{appx:implementationdetail}

Our implementation of \name~is based on PyMARL~\footnote{We use PyMARL with SC2.4.6.2.6923 for the experiments on SMAC-1o\_2r\_vs\_4r and SMAC-1o\_10b\_vs\_1b. Note that performance is not always comparable among versions} with StarCraft 2.4.6.2.69232. We adopt its default settings for some common hyper-parameters like learning rate. The choices of other hyper-parameters in our experiments are listed in Tab.~\ref{tab:hyper-parameter}.

\begin{table*}[t]
\centering
\caption{Selected hyper-parameters in our experiments.}
\label{tab:hyper-parameter}
\begin{tabular}{C{4cm}|c|c|c|c}
\toprule
Hyper-parameter Name & Other Experiments & Hallway-4x5x9 & TarMAC: TJ & GP-4r, GP-9r \\
\midrule
Population Size (The number of attackers one population contains)                                                         & \multicolumn{4}{c}{20}     \\\midrule
Reproduction Ratio (The proportion of updated agents in the population during each evolution)                             & \multicolumn{4}{c}{0.25}   \\\midrule
Distance Threshold (The threshold to determine whether the new attacker is novel enough)                                  & \multicolumn{4}{c}{0.25}   \\\midrule
Critic Update Times (The number of times the critic is updated for each actor update in MATD3)                            & \multicolumn{4}{c}{5}      \\\midrule
Num of Sampled Trajectories (The number of sampled trajectories used to encode attacker identification)                   & \multicolumn{4}{c}{10}     \\\midrule
Alternate Update Times (The number of iterations of alternate updates between the ego system and the attacker population) & 15 & 30 & 20 & 6     \\\midrule
Num of Samples for Ego System in One Loop (The number of samples used to update the ego system in each iteration)         & \multicolumn{3}{c}{205000} \vline & 505000 \\\midrule
Evolution Times in One Loop (The number of times the population conduct evolution in each iteration)                      & \multicolumn{4}{c}{10}     \\\midrule
Num of Samples for Attacker in One Evolution (The number of samples used to update the attacker in each evolution)        & \multicolumn{4}{c}{10000}  \\
\bottomrule
\end{tabular}
\end{table*}

\subsection{Experimental Details} \label{appx:experimentaldetail}
In this section, we provide more experimental details to help the reader better understand or reproduce our experimental results. Specifically, we discuss the details about robustness comparison (Sec.~\ref{exp:main_performance}), attacking behavior analysis (Sec.~\ref{exp:behavior_analysis}), policy transfer (Sec.~\ref{exp:transfer}), and parameter sensitivity studies (Sec.~\ref{exp:parameter_study}), respectively. Note that all experiments are conducted with five independent runs and we report the mean and standard deviation. 
\paragraph{Robustness Comparison} As we have mentioned in the section of Problem Setting, if perturbation without bounds is allowed, the attacker can be arbitrarily strong and effective defence may be impossible. Thus, to consider a more realistic setting, we specify that the perturbed messages are restricted in a specific set, e.g., the perturbed messages $\hat{m}$ ought to satisfy $\|m-\hat{m}\|_p \le \epsilon$ with respect to the original messages $m$. Actually, we adopted a practice that the perturbed messages are constrained to a 1-norm ball centered on the original communication messages, which means the set $\mathcal{B}=\{\hat{m} \mid \|m-\hat{m}\|_1 = \epsilon \}$ for most experiments, except for the experiments on Hallway. The observation dimension on Hallway equals to $1$; thus if we apply the previous practice, the action space will degenerate to a discrete action space of $\{\epsilon, -\epsilon\}$ when attacking the communication system learned with Full-Comm. For experiments on Hallway, we constrain $\hat{m}$ to satisfy $\|m -\hat{m}\|_1 \le \epsilon$. Besides, the adopted magnitude $\epsilon$ for each experiment is listed in Tab.~\ref{tab:magnitude}.
\begin{table}[htbp]
\caption{Adopted magnitude $\epsilon$ for each experiment. Comm. Alg. is short for Communication Algorithm.}
\label{tab:magnitude}
\centering
    \begin{tabular}{ccccc}
    \toprule
    Comm. Alg.   & \multicolumn{4}{c}{Full-Comm}         \\
    \cmidrule(lr){1-1}
    \cmidrule(lr){2-5}
    Task       & Hallway-6x6 & Hallway-4x5x9 & SMAC-1o\_2r\_vs\_4r & SMAC-1o\_10b\_vs\_1r \\
    \cmidrule(lr){1-1}
    \cmidrule(lr){2-5} 
    $\epsilon$ & 1.5         & 1.0           & 10               & 25                \\
    \midrule
    Comm. Alg.   & \multicolumn{2}{c}{Full-Comm}  & NDQ              & TarMAC            \\
    \cmidrule(lr){1-1}
    \cmidrule(lr){2-3}
    \cmidrule(lr){4-4}
    \cmidrule(lr){5-5}
    Task         & GP-4r       & GP-9r         & SMAC-1o\_2r\_vs\_4r & TJ                \\
    \cmidrule(lr){1-1}
    \cmidrule(lr){2-3}
    \cmidrule(lr){4-4}
    \cmidrule(lr){5-5}
    $\epsilon$   & 2           & 2             & 6                & 16                \\
    \bottomrule
    \end{tabular}
\end{table}

Besides, for the test mode of \textbf{Aggressive Attackers}, we additionally train a set of unseen communication attackers and utilize them to test the robustness of different methods. Specifically, we extraly train a diverse attacker population with the technique of \name. Then we evenly split the training process of the population into five stages, and randomly sample four attacker models from the stored models at each stage, finally resulting in a total of $20$ attacker models. We adopt the average performance of the ego system policy under the attacks of these $20$ attacker models to represent the robustness performance.
\paragraph{Attacking Behavior Analysis} In the experiments for Attacking Behavior Analysis, we studied the attack ability of our approach and the obtained attacker population. Specifically, for both the experiments of attack ability comparison and population visualization, we firstly pre-train a communication-based policy for the ego system, then we apply MATD3 and TD3 to learn attacker models to this ego system for the experiments of attack ability comparison, and apply \name~to optimize an attacker population for population visualization. Besides, the adopted magnitude $\epsilon$ is consistent with that in the section of Robustness Comparison, $10$ for SMAC-1o\_2r\_vs\_4r and $2$ for GP-4r.
\paragraph{Policy Transfer} In this part, we investigated the generalization ability of our approach to different perturbation ranges. Actually, we utilize the same setting as before, which means that we constrain the perturbed messages to be on a 1-norm ball centered on the original messages on task of SMAC and GP. However, we modify the magnitude $\epsilon$ to test the generalization ability and transfer ability of our approach. For the zero-shot generalization test, we select a magnitude set of $\{5, 10, 20, 40, 60, 80\}$ for SMAC-1o\_2r\_vs\_4r, where $10$ is the magnitude for adversarial training, and select a magnitude set of $\{1, 2, 4, 6, 8\}$ for GP-4r, where $2$ is the magnitude for adversarial training. For the fine-tuning transfer test, the magnitudes for training are $1$ and $2$ respectively for Hallway-4x5x9 and GP-4r, while the transfer perturbation ranges are respectively $5$ and $15$.
\paragraph{Parameter Sensitivity Studies} In the section of Parameter Sensitivity Studies, we selectively experiment on the task of GP-4r to investigate the parameter sensitivity of our approach. In specific, we consider three main hyper-parameters, which are Population Size, Reproduction Ratio, and Distance Threshold, respectively. When studying the influence of one specific hyper-parameters, the other hyper-parameters are set as the default values utilized in the main experiments. For example, when we investigate the hyper-parameters Population Size, the Reproduction Ratio and Distance Threshold are both set as $0.25$. The other settings like perturbation range are the same as those in the experiments of Robustness Comparison.

\subsection{Additional experiments for the AME baseline}\label{appx:extra_experiment}
\begin{table}[t!]
\centering
\caption{Additional test results for the AME baseline.}\label{tab:ame}
\begin{tabular}{C{4cm}C{2.4cm}C{2cm}C{2cm}}
\toprule
                     &                              & MA3C             & AME   \\
\midrule
\multirow{3}{*}{SMAC-1o\_2r\_vs\_4r} & Normal            & {0.86$\pm$0.02}           & {0.81$\pm$0.05}  \\
                     & Random Noise                      & {0.84$\pm$0.02}           & {0.76$\pm$0.07}  \\
                     & Aggressive Attackers              & {0.81$\pm$0.01}           & {0.60$\pm$0.06}  \\
\cline{1-4}
\multirow{3}{*}{GP-4r}   & Normal                & {0.87$\pm$0.02}           & {0.23$\pm$0.37}  \\
                     & Random Noise                      & {0.87$\pm$0.02}           & {0.24$\pm$0.40}  \\
                     & Aggressive Attackers              & {0.86$\pm$0.01}           & {0.17$\pm$0.29}  \\
\bottomrule
\end{tabular}
\end{table}
\add{Considering that the AME approach is more suited to the test setting where a part of communication channels are attacked, we further conduct the test in this setup to justify the effectiveness of our approach. Specifically, we conduct experiments in the task of SMAC-1o\_2r\_vs\_4r and GP-4r, and the results are reported in Tab.~\ref{tab:ame}. From the results, we can see that the AME approach shows better robustness when only partial communication channels are attacked compared with the results in Tab.~\ref{tab:main_exp}. However, our approach MA3C still exhibits good performance advantages over the AME baseline, which validates the effectiveness of our approach.}

\bibliographystyle{fcs}
\bibliography{sample}

\end{document}